\def\cvprPaperID{736}
\def\confName{CVPR}
\def\confYear{2023}

\def\paperTitle{Advancing Visual Grounding with Scene Knowledge: Benchmark and Method}

\def\authorBlock{
    Zhihong Chen$^{1,2,3}$\footnotemark[1] \qquad
    Ruifei Zhang$^{2}$\footnotemark[1] \qquad
    Yibing Song$^{4,5}$ \qquad
    Xiang Wan$^{3}$ \qquad 
    Guanbin Li$^{2}$\footnotemark[2]\\
    $^{1}$The Chinese University of Hong Kong, Shenzhen \quad $^{2}$Sun Yat-sen University \\
    $^{3}$Shenzhen Research Institute of Big Data \quad $^{4}$Tencent AI Lab \quad $^{5}$AI$^3$ Institute, Fudan University\\
    {\tt\small zhihongchen@link.cuhk.edu.cn ~ zhangrf23@mail2.sysu.edu.cn} \\
    {\tt\small yibingsong.cv@gmail.com ~ wanxiang@sribd.com ~ liguanbin@mail.sysu.edu.cn}
}

\newif\ifreview 
\newif\ifarxiv 
\newif\ifcamera \newcommand{\cameraready}{\cameratrue}
\newif\ifrebuttal

\cameraready % \review OR \arxiv OR \cameraready

\pdfoutput=1
\documentclass[10pt,twocolumn,letterpaper]{article}
\ifreview \usepackage[review]{cvpr} \fi
\ifarxiv \usepackage[pagenumbers]{cvpr} \fi
\ifrebuttal \usepackage[rebuttal]{cvpr} \fi
\ifcamera \usepackage{cvpr} \fi

\usepackage{graphicx}
\usepackage{amsmath}
\usepackage{amssymb}
\usepackage{booktabs}

\usepackage{times}
\usepackage{microtype}
\usepackage{epsfig}
\usepackage[table,xcdraw]{xcolor}
\usepackage{caption}
\usepackage{float}
\usepackage{placeins}
\usepackage{color, colortbl}
\usepackage{stfloats}
\usepackage{enumitem}
\usepackage{tabularx}
\usepackage{xstring}
\usepackage{multirow}
\usepackage{xspace}
\usepackage{url}
\usepackage{subcaption}
\usepackage{xcolor}
\usepackage[hang,flushmargin]{footmisc}

\ifcamera \usepackage[accsupp]{axessibility} \fi

% ...

\ifarxiv  \fi

\newcommand{\R}[1]{{%
    \textbf{%
        \ifstrequal{#1}{1}{\textcolor{red}{R#1}}{%
        \ifstrequal{#1}{2}{\textcolor{blue}{R#1}}{%
        \ifstrequal{#1}{3}{\textcolor{magenta}{R#1}}{%
        \ifstrequal{#1}{4}{\textcolor{teal}{R#1}}{%
                           \textcolor{cyan}{R#1}%
        }}}}%
    }%
}}

\usepackage{xr-hyper}

\makeatletter
\newcommand*{\addFileDependency}[1]{
  \typeout{(#1)}
  \@addtofilelist{#1}
  \IfFileExists{#1}{}{\typeout{No file #1.}}
}

\makeatother

\usepackage[table]{xcolor}
\usepackage[pagebackref, breaklinks=true, colorlinks, citecolor=citecolor, linkcolor=linkcolor, bookmarks=false]{hyperref}
\definecolor{citecolor}{HTML}{0071BC}
\definecolor{linkcolor}{HTML}{ED1C24}
\usepackage[capitalize]{cleveref}
\crefname{section}{Sec.}{Secs.}
\crefname{table}{Table}{Tables}
\crefname{figure}{Fig.}{Figs.}

\begin{document}
%% TITLE
\title{\paperTitle}
\author{\authorBlock}
\maketitle
\def\thefootnote{*}\footnotetext{Equal contribution}
\def\thefootnote{\dag}\footnotetext{Corresponding author}
\def\thefootnote{\arabic{footnote}}

\begin{abstract}
Visual grounding (VG) aims to establish fine-grained alignment between vision and language.
Ideally, it can be a testbed for vision-and-language models to evaluate their understanding of the images and texts and their reasoning abilities over their joint space.
However, most existing VG datasets are constructed using simple description texts, which do not require sufficient reasoning over the images and texts.
This has been demonstrated in a recent study~\cite{luo2022goes}, where a simple LSTM-based text encoder without pretraining can achieve state-of-the-art performance on mainstream VG datasets.
Therefore, in this paper, we propose a novel benchmark of \underline{S}cene \underline{K}nowledge-guided \underline{V}isual \underline{G}rounding (SK-VG), where the image content and referring expressions are not sufficient to ground the target objects, forcing the models to have a reasoning ability on the long-form scene knowledge.
To perform this task, we propose two approaches to accept the triple-type input, where
the former embeds knowledge into the image features before the image-query interaction;
the latter leverages linguistic structure to assist in computing the image-text matching.
We conduct extensive experiments to analyze the above methods and show that the proposed approaches achieve promising results but still leave room for improvement, including performance and interpretability.
The dataset and code are available at \url{https://github.com/zhjohnchan/SK-VG}.
\end{abstract}

\section{Introduction}
\label{sec:intro}
% ****************** Figure 1 ******************
\begin{figure}[t]
\centering
\includegraphics[width=0.45\textwidth, trim=0 0 0 0]{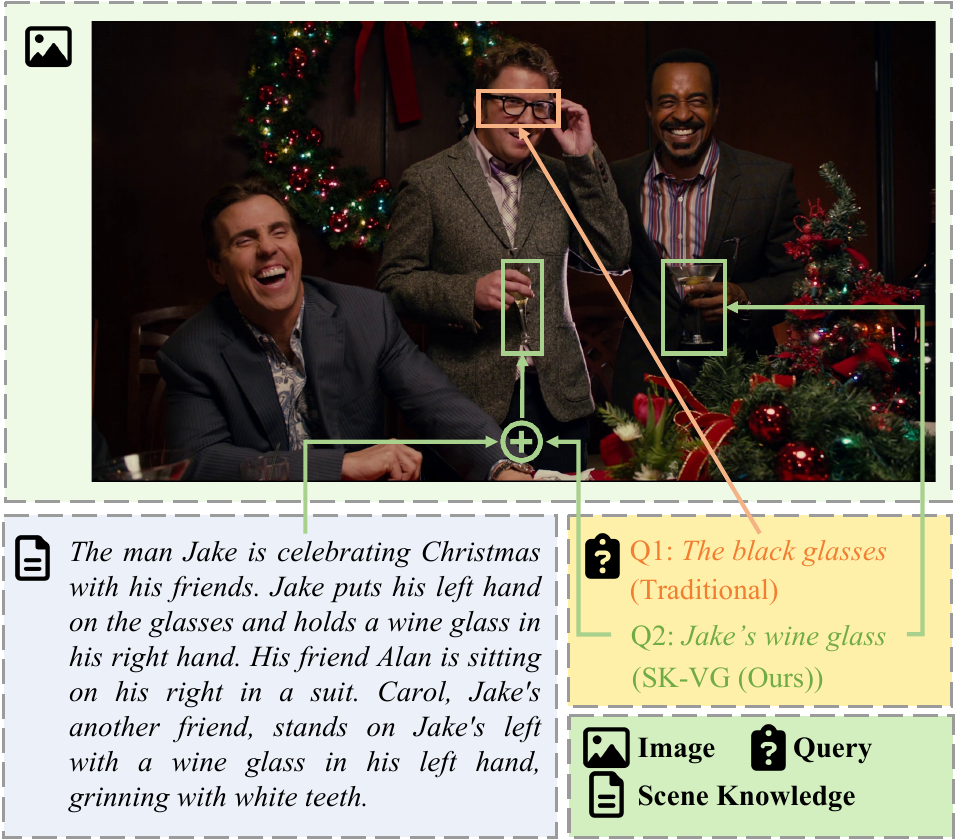}
\caption{An example from the proposed SK-VG dataset for scene knowledge-guided visual grounding.
The task requires a model to reason over the (image, scene knowledge, query) triple to locate the target object referred to by the query.}
\label{fig:example}
\end{figure}
% ****************** Figure 1 ******************
Visual grounding (VG), aiming to locate an object referred to by a description phrase/text in an image, has emerged as a prominent attractive research direction.
It can be applied to various tasks (e.g., visual question answering \cite{zhu2016visual7w,gan2017vqs,wang2020general,chen2021grounding} and vision-and-language navigation \cite{anderson2018vision,vogel2010learning,ding2022embodied}) and also be treated as a proxy to evaluate machines for open-ended scene recognition.
Typically, VG requires models to reason over vision and language and build connections through single-modal understanding and cross-modal matching.
Yet, current VG benchmarks (e.g., RefCOCO \cite{yu2016modeling}, RefCOCO+ \cite{yu2016modeling}, RefCOCOg \cite{mao2016generation}, ReferItGame \cite{kazemzadeh2014referitgame}, and CLEVR-Ref+ \cite{liu2019clevr}) can not serve as a good test bed to evaluate the reasoning ability since they only focus on simple vision-language alignment.
In addition to the simple nature of constructed referring expressions, this can be reflected in the recent state-of-the-art study \cite{luo2022goes}, where they showed that \textit{VG models are less affected by language modeling through extensive empirical analyses}.

% ****************** Figure 1 ******************
\begin{figure*}[t]
\centering
\includegraphics[width=0.95\textwidth, trim=0 0 0 0]{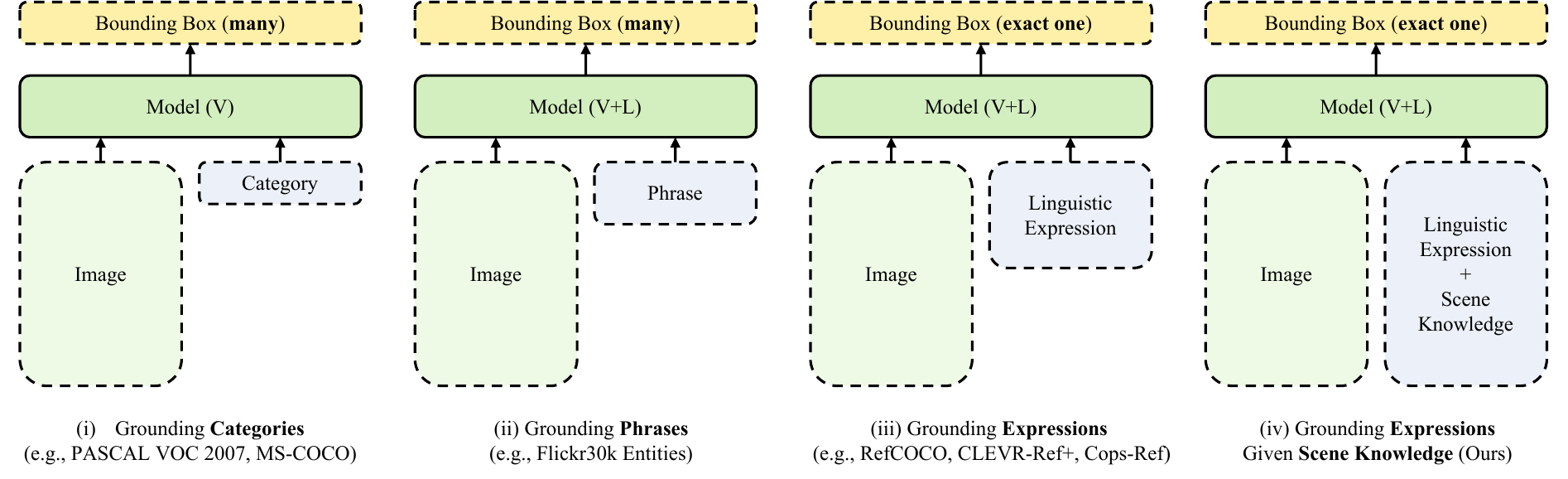}
\caption{Illustrations of four categories of grounding tasks, including categories, phrases, linguistic expressions, and linguistic expression+scene knowledge.
The height of the input green and blue rectangles denotes its relative information.}
\label{fig:taxonomy}
\end{figure*}
% ****************** Figure 1 ******************

In this paper, we believe that the intrinsic difficulty of VG lies in the difference between perceptual representations of images and cognitive representations of texts.
Specifically, visual features are obtained through perceptual learning, which only maps visual appearances in images to semantic concepts.
However, open-ended queries might require VG models to understand the whole scene knowledge before performing reasoning to locate the target object.
As shown in Figure \ref{fig:example}, the perceptual features can encode the information about ``\textit{a wine glass}'', but it would struggle to locate ``\textit{Jake’s wine glass}'' without the scene knowledge about ``\textit{who is Jake?}''.
This is a challenging task owing to two facts:
(i) From the dataset perspective, there are no relevant benchmarks for the VG researchers to evaluate their models;
(ii) From the model/algorithm perspective, it is not easy to design models to perform reasoning among images, scene knowledge, and open-ended querying texts.

Therefore, we propose to break this limitation of current VG research and construct a new benchmark requiring \underline{VG} to perform reasoning over \underline{S}cene \underline{K}nowledge (i.e., text-based stories).
The benchmark named SK-VG contains $\sim$40,000 referring expressions and 8,000 scene stories from 4,000 images, where each image contains 2 scene stories with 5 referring expressions for each story.
Moreover, to evaluate the difficulty levels of queries, we curate the test set by splitting the samples into easy/medium/hard categories to provide a detailed evaluation of the vision-language models.
Under this new setting, we develop a one-stage approach (i.e., Knowledge-embedded Vision-Language Interaction ({KeViLI})) and a two-stage approach (i.e., Linguistic-enhanced Vision-Language Matching ({LeViLM})).
In KeViLI, the scene knowledge is firstly embedded into the image features, and then the interaction between the image and the query is performed;
In LeViLM, the image features and the text features are first extracted, and then the matching between the (image) regions and the (text) entities are computed, assisted by the structured linguistic information.
Through extensive experiments, we show that the proposed approaches can achieve the best performance but still leave room for improvement, especially in the hard split.
It challenges the models from three perspectives: First, it is an open-ended grounding task; Second, the scene stories are long narratives consisting of multiple sentences; Third, it might require the multi-hop reasoning ability of the models.
In summary, the contributions of this paper are three-fold:
\begin{itemize}[noitemsep,nolistsep]
\item We introduce a challenging task that requires VG models to reason over (image, scene knowledge, query) triples and build a new dataset named SK-VG on top of real images through manual annotations.
\item We propose two approaches to enhance the reasoning in SK-VG, i.e., one one-stage approach KeViLI and one two-stage approach LeViLM.
\item Extensive experiments demonstrate the effectiveness of the proposed approaches. Further analyses and discussions could be a good starting point for future study in the vision-and-language field.
\end{itemize}

\section{Background}
\label{sec:related}
\subsection{Taxonomy of Visual Grounding Datasets}
In the past few years, a variety of datasets have been proposed for visual grounding.
We propose a taxonomy of existing (generalized) VG datasets along with the proposed dataset based on types of queries, as shown in Figure~\ref{fig:taxonomy}.

The datasets of the first type use \textcolor{magenta}{fixed} categories as queries.\footnote{Generally, it is called object detection. We generalize it to visual grounding to summarize and classify existing grounding datasets better.}
Grounding categories in images are a fundamental task in computer vision and has attracted much attention.
One of the most representative examples is the MS-COCO dataset~\cite{lin2014mscoco}, which contains 80 categories.
Besides, PASCAL VOC 2007~\cite{everingham2010pascal}, Visual Genome~\cite{krishna2017vg}, and Object365~\cite{shao2019objects365} are also popular datasets of this type.

The Flickr30K Entities dataset\footnote{Although there are some studies locating the entities using the context provided by the dataset, we refer in particular to those using only the phrases as in \cite{luo2022goes}.} \cite{plummer2015flickr30k} belongs to the second type, where the queries are short phrases.
Similar to the first type, an image might contain multiple objects referred to by a phrase following the \textcolor{cyan}{one-to-many} mappings.
The most distinct characteristic of this type from the first type is that it is an \textcolor{magenta}{open-vocabulary} grounding problem instead of using fixed categories.
Most recently, researchers constructed relevant datasets of this type, i.e., PhraseCut~\cite{wu2020phrasecut} and LVIS~\cite{gupta2019lvis}, with a larger scale.

The third type aims at localizing a specific object in the image based on an expression in the form of natural language.
In the narrow sense, the term \textit{visual grounding} refers to this type of dataset in previous studies.
Various benchmark datasets (e.g., RefCOCO~\cite{yu2016modeling}, RefCOCO+~\cite{yu2016modeling}, RefCOCOg~\cite{mao2016generation}, and CLEVR-Ref+~\cite{liu2019clevr}) have been constructed to test the ability to refer expression comprehension of existing vision-language models.
In general, expressions in these datasets are written according to the visual appearance and spatial location of an object, where the visual appearance includes visual categories, color, and other visual attributes, and the spatial location describes the absolute or relative location.
Different from the aforementioned two types, an expression in this type of dataset points to a unique object in the image following the \textcolor{cyan}{one-to-one} mapping.

Our proposed SK-VG is the first dataset of the fourth type, where for each image, we provide human-written scene knowledge to describe its content.
By doing so, the VG models need to have a good understanding of the scene stories and then locate the queried object in the image according to both querying expressions and scene stories.
Although there exists a dataset~\cite{wang2020give} introducing knowledge to the visual grounding model, it only focuses on the commonsense knowledge, which interprets the concept in the referring expressions, e.g., the interpretation of the target object `banana'.
There are also some datasets on grounding complex/compositional visual description, e.g., the human-centric HumanCog dataset~\cite{you2022humancog} and the Cops-Ref dataset~\cite{chen2020cops}. HumanCog requires the model to understand human-centric commonsense (e.g., the mental aspect), and Cops-Ref proposed a difficult task to require a model to identify an object described by a compositional referring expression from a curated set of images.
We can still classify them into the first three categories since the knowledge is more about referring expressions, while the knowledge in our dataset is a comprehensive description of the scene.

% ******* Figure ********
\begin{figure*}[t]
\centering
\includegraphics[width=0.93\textwidth, trim=0 0 0 0]{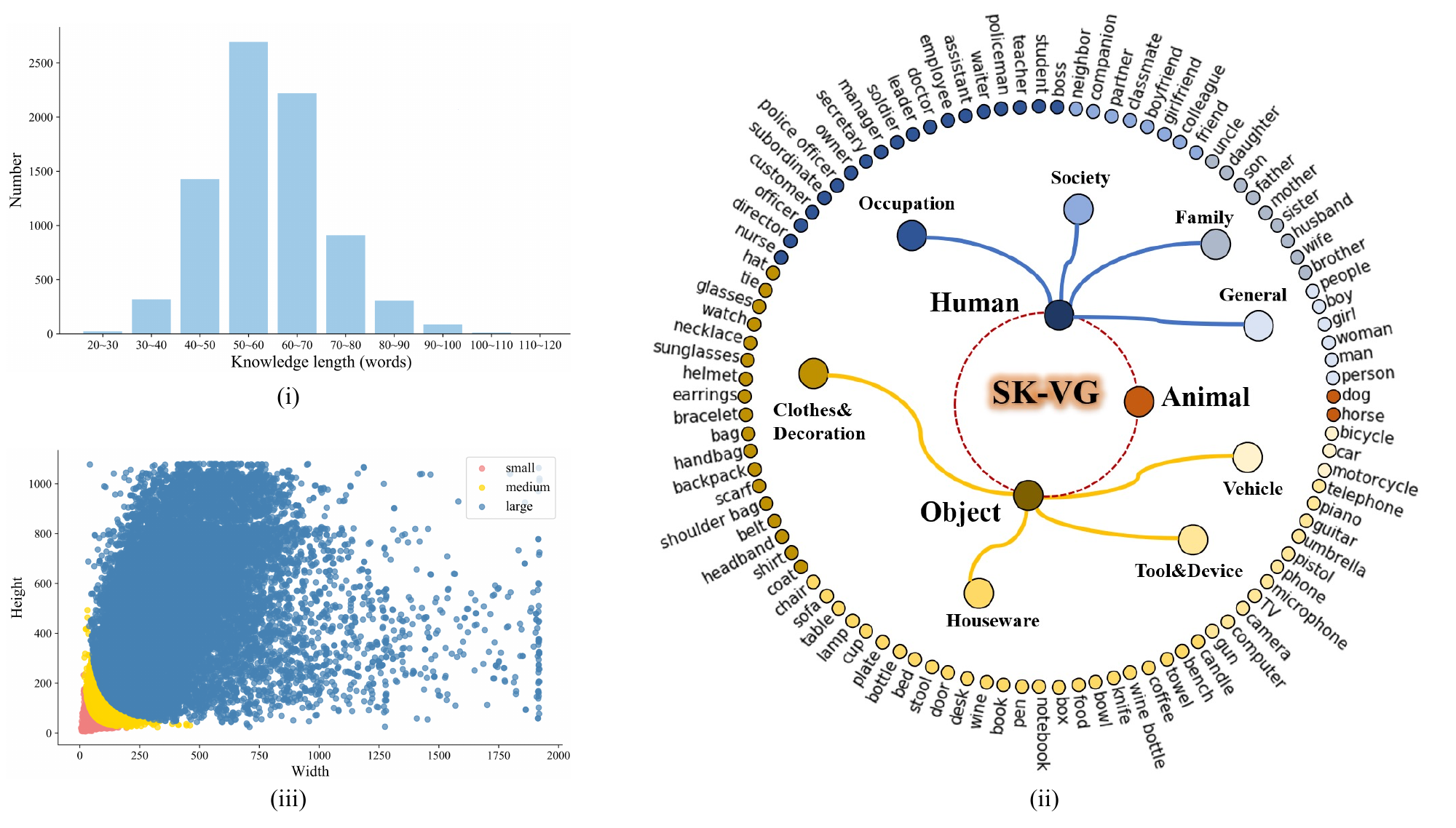}
\caption{Statistics of the proposed SK-VG dataset: (i) the length distribution of the knowledge description; (ii) the referred objects of high-frequency; (iii) the size distribution of referred objects.}
\label{fig:dataset_stat}
\end{figure*}
% ******* Figure ********

\subsection{Visual Grounding Models}
Existing methods can be categorized into two classes: (i) two-stage methods~\cite{yu2018mattnet,wang2019neighbourhood,liu2019learning,bajaj2019g3raphground,liu2019improving} and (ii) one-stage methods~\cite{yang2019fast,luo2020multi,yang2020improving,zhou2021real,deng2021transvg,huang2021look}.\footnote{Grounding categories is a very hot topic, where there are many research works~\cite{ren2015faster,liu2016ssd,redmon2016you,liu2020deep}. Yet we mainly discuss existing studies of grounding phrases/expressions, which are more related to this work.}
The former generates region proposals first and then exploits the language expression to select the best-matching region;
The latter directly predicts the bounding boxes through vision-and-language interaction to avoid the computation-intensive object proposal generation and region feature extraction in the two-stage paradigm.
Among these methods, some work~\cite{wang2019neighbourhood,yang2019dynamic,yang2020graph,chencomphy,ding2021dynamic} perform explicit reasoning by modeling the attributes of objects and the relations between objects to improve interpretability.
However, limited by the simplicity of existing datasets, they can not take full advantage of their algorithms and do not model complicated semantic relations in images and texts.
Besides, pretraining-based methods~\cite{kamath2021mdetr,li2022grounded,zhang2022glipv2,yang2022unitab} have been applied to VG to improve the open-vocabulary grounding ability.

\section{Dataset Construction}
\label{sec:dataset}
In this section, we present the SK-VG dataset.
Compared to existing VG tasks, the key difference is that each image is paired with scene knowledge to describe its content.
We detail the image collection, the annotation process, the dataset statistics, and splits in the following subsections.

\subsection{Image Collection}
To facilitate and ease the writing of a text story, we identify three significant aspects a qualified image should fulfill:
\begin{itemize}[noitemsep,nolistsep]
\item \textbf{Humans} are the main body of a story. A satisfying image is better full of multiple characters with interactions to create complex and dramatic stories.
\item \textbf{Objects} are also essential and necessary to complement the details of a story. The number and category of objects also impact our SK-VG task, making it more challenging and interesting.
\item \textbf{Scenes} are the third factors we can not ignore since the scenes determine the background and starting point of stories. Complex and real scenes (e.g., theaters, classrooms, and parks) can inspire diverse stories. 
\end{itemize}
Based on the above consideration, we select the existing Visual Commonsense Reasoning dataset \cite{zellers2019recognition}, which is designed for the visual question answering task and contains more than 110,000 movie scene images.
Thanks to the movie attribute of these images, they are more likely to meet our requirements and are suitable for our task.
Therefore, through careful manual filtering and selection, 4,000 images serve as the ingredients of our SK-VG dataset.

\subsection{Image Annotation}
To facilitate the annotation process, we develop software.
For each image, the annotation mainly includes two phases: (i) Annotators are asked to create two different story descriptions based on a given image; (ii) Given each story, the annotators are asked to write five referring expressions related to the given image and story and annotate the corresponding object bounding boxes as the ground truth.
The required rules for each step are detailed as follows:
(i) Knowledge Annotation serves as a foundation stone of our annotation, determining the scope and quality of query sentences.
A satisfying story should be related but beyond the image content.
Specifically, the story should cover the person who occurred in one image with accurate visual descriptions, thus providing significant clues and evidence to match the image object and knowledge entity.
Besides, the story is required to contain more context beyond the image, such as background, character relationship, mental state, and emotion, so as to promote the design of more challenging and flexible query expression.
(ii) Query Expression Annotation plays an essential role in our task and ought to obey the following criteria:
\begin{itemize}[noitemsep,nolistsep]
\item \textbf{Knowledge Relevance:}
The main insight of our SK-VG task is to advance the traditional VG task by introducing extra scene knowledge descriptions. Based on this consideration and prospect, the first principle is that query sentences must be highly relevant to knowledge instead of directly visually distinguishable. Taking Figure~\ref{fig:example} as an example, ``\textit{The black glasses}'' is not qualified since it does not involve knowledge information.
\item \textbf{Uniqueness:}
To give a unique bounding box of the referred object, the query should be clear and unambiguous. For instance, queries like ``\textit{The person holding a wine glass}'' or ``\textit{Jake's friend}'' are not satisfied in Figure~\ref{fig:example} since they involve several objects in the image.
\item \textbf{Diversity:}
For one thing, the referring objects should be diverse; For another, the lexical expression of the query sentence is also required to be diversified. For example, the general terms (e.g., ``\textit{person}'') could be replaced by other specific alternatives (e.g., ``\textit{colleague}'').
\end{itemize}

\subsection{Dataset Statistics}
To further dive into the proposed SK-VG dataset, we demonstrate its characteristics from three aspects:
\begin{itemize}[noitemsep,nolistsep]
\item \textbf{Length of scene knowledge}:
As shown in Figure \ref{fig:dataset_stat}(i), the word-based length of most stories ranges from 50 to 70.
This puts high demands on models to capture long-range dependency to understand text content.
\item \textbf{Categories of referred objects}:
As an open-world task, referred objects of our dataset are not limited to a fixed number of categories. Figure \ref{fig:dataset_stat}(ii) exhibits 100 referred object classes with the highest frequency. Benefiting from the diverse stories and scenes, we introduce extensive referred targets with various expressions, increasing the difficulty of recognition and localization.
\item \textbf{Size of referred objects}:
We report the size of referred objects in Figure \ref{fig:dataset_stat}(iii), which indicates that the objects in our dataset fall into a wide range of sizes. Further, we define small, medium, and large concepts according to the area of the objects, following the boundary of $64\times64$ and $128\times128$. We can observe that large objects dominate our dataset, while small and medium instances hold a small proportion.
\end{itemize}

% ****************** Figure 2 ******************
\begin{figure*}[t]
\centering
\includegraphics[width=0.9\textwidth, trim=0 10 0 0]{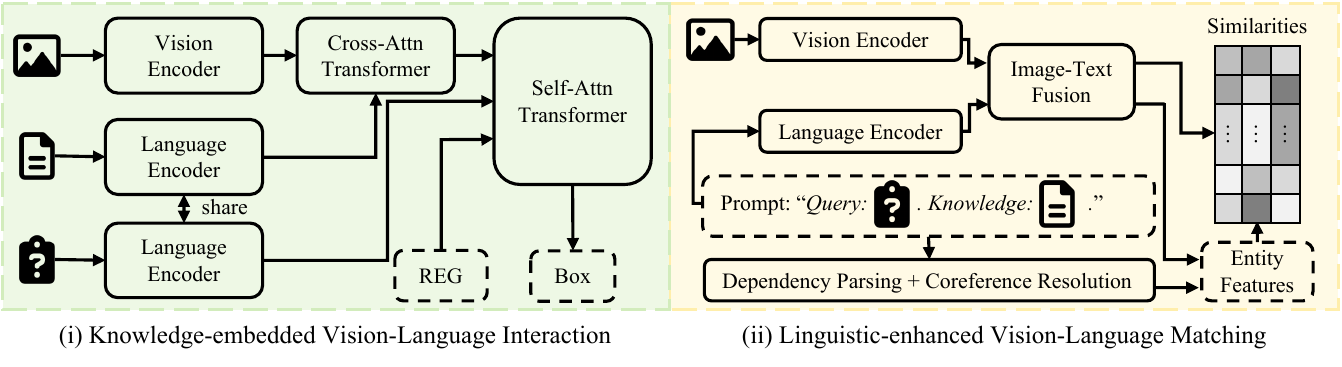}
\caption{Illustration of the proposed approaches: (i) the one-stage algorithm, where the knowledge is embedded into the image features before the image-query interaction; (ii) the two-stage algorithm, where the image features and text features are firstly extracted, and then the structured linguistic information is leveraged to assist in computing the region-entity similarities.}
\label{fig:model-architecture}
% \vskip -1em
\end{figure*}
% ****************** Figure 2 ******************

\subsection{Dataset Splits}
We randomly sample 60\% of images and their annotations as the training set.
For the remaining (image, scene knowledge, query) triples, we sample parts of them for annotating their difficulty levels and use them as the test set while the remaining triples are used as the validation set.\footnote{The images in the training set have no overlap with those in the validation and test sets.}
We follow the following rule to annotate the difficulty level.
The core principle is that more knowledge-related but less visual-distinguishable expressions deserve a higher difficulty level: (i) Easy: The referring expression contains obvious appearance, object relationship, or other visual clues; (ii) Medium: The expression only mentions weak visual information; (iii) Hard: The answer is required to be entirely derived from the scene knowledge without visual bias.
We show examples of different difficulty levels in \S\ref{subsec:case_study}.

\section{Algorithmic Analysis}
\subsection{Algorithm 1: KeViLI}
To perform SK-VG, we introduce a one-stage algorithm: Knowledge-embedded Vision-Language Interaction ({KeViLI}).
Given an image $I$ and its corresponding scene knowledge $K$, the goal is to locate the object referred to by a querying text $T$ by predicting the coordinates of its corresponding bounding box directly.

In detail, given $I$, we use an image encoder to encode $I$ to the image patch features $H_{I}$; given $K$ and $T$, we use a language encoder to encode them to the knowledge features $H_{K}$ and the text subword features $H_{T}$, respectively.
Afterward, we embed the scene knowledge into the image features before the image-query interaction.
As an intuitive illustration, in Figure~\ref{fig:example}, the visual features of ``\textit{person}'' in the image is not only about the concept ``\textit{person}'' but also about the specific refer ``\textit{Jake}'' after embedding knowledge, which can assist in grounding ``\textit{Jake's wine glasses}''.
The embedding procedure is implemented using a cross-attention Transformer, which is stacked by self-attention, cross-attention, and feed-forward sub-layers.
The attention mechanism is applied in the self-attention and cross-attention sub-layers and is defined as
\begin{equation}
    \text{ATTN}(\boldsymbol{Q}, \boldsymbol{K}, \boldsymbol{V})=\operatorname{Softmax}\left(\boldsymbol{Q} \boldsymbol{K}^{\top}/\sqrt{D_{k}}\right) \cdot \boldsymbol{V},
\end{equation}
In the self-attention sub-layer, the patch features interact with each other through $H^{I}=\text{ATTN}(H^{I}, H^{I}, H^{I})$;
In the cross-attention sub-layer, the knowledge is embedded into the image features by $H^{I}=\text{ATTN}(H^{I}, H^{K}, H^{K})$.\footnote{We overload the notations here for simplicity.}
Subsequently, $H_{I}$ and $H_{T}$ are input to a Transformer with a learnable regression token \texttt{[REG]} to perform the image-query interaction.
The output of \texttt{[REG]} is input to a two-layer multilayer perceptron (MLP) to produce to predict the coordinates directly without region proposals.
The model is trained to minimize the generalized IoU loss \cite{rezatofighi2019generalized} (GIoU loss):
\begin{equation}
    L = L_{\text{smooth\_l1}}(b, \hat{b}) + L_{\text{giou}}(b, \hat{b}),
\end{equation}
where $b$ and $\hat{b}$ refer to the ground-truth and prediction boxes, respectively and $L_{\text{smooth\_l1}}(\cdot)$ and $L_{\text{giou}}(\cdot)$ are the smooth L1 loss and GIoU loss, respectively.

\subsection{Algorithm 2: LeViLM}
We further introduce a two-stage algorithm: Linguistic-enhanced Vision-Language Matching ({LeViLM}).
In LeViLM, we follow GLIP~\cite{li2022grounded} to initialize the backbone model, which had been trained from large-scale datasets to detect objects of open-vocabulary classes.

In detail, the grounding process is disentangled into two stages, i.e., region proposal and scoring, where the former aims to find all the objects in the image, and the latter aims to score the proposal regions.
In the region proposal stage, given the scene knowledge $K$ and the query $T$, we construct a manual prompt text $P$: ``\texttt{Query: $T$. Knowledge: $K$.}''.
Then we use a language encoder to encode $P$ into the prompt features $H_P$ and an image encoder to encode $I$ to the image features $H_I$.
Afterward,  we perform the image-text fusion using a stack of $L$ layers.
In each layer, there is one self-attention layer for text encoding and one Dynamic Head layer~\cite{dai2021dynamic} for image encoding, and two cross-attention layers for cross-modal fusion.
The text encoding and image encoding process can be formulated as $H_P = \operatorname{ATTN}(H_P, H_P, H_P)$ and $H_I = \operatorname{Dynamic Head}(H_I)$, respectively.
The cross-modal information fusion process can be formalized as $H_P = \operatorname{ATTN}(H_P, H_I, H_I), H_I = \operatorname{ATTN}(H_I, H_P, H_P)$, where the cross-attention mechanism is applied to exchange the image and text information.
Subsequently, a region proposal layer is applied to $H_I$ to obtain the region features.
For simplicity, we denote the after-fusion image (region) features and text (subword) features as $Z_I \in \mathrm{R}^{N \times d}$ and $Z_P \in \mathrm{R}^{M \times d}$, where $N$ refers to the number of proposed regions and $M$ represents the number of subwords.
In the region scoring stage, we extract structured linguistic information from the query $T$ and the scene knowledge $K$.
Specifically, given $T$, we perform syntactic dependency parsing to obtain its dependency tree and apply a set of rules to extract the subject of $T$, which we denote as the head entity $E_h$.
Besides, we also build the connection between $T$ and $K$ through coreference resolution to find all mentions $E_m$ in $K$ refer to the same underlying entity $E_h$.
Therefore, during the training procedure, we have the bounding box annotation for $E_h$ and its co-referred $E_m$ since $E_h$ and $E_m$ share the same object.
Then we can take the representations of $E_h$ and $E_m$ from $Z_P$, denoted as $Z_E \in \mathrm{R}^{(E+1) \times d}$, where $E$ represents the number of co-referred mentions.
Afterward, we can compute the alignment scores between the image regions and the entities in the prompt:
\begin{equation}
    Score = Z_{I}Z_{e}^{\top},
\end{equation}
where $Score \in \mathrm{R}^{N \times (E+1)}$.
Finally, the model is trained to minimize the following loss:
\begin{equation}
    \label{equation:ce_loss}
    L = L_{xe}(Score, Target),
\end{equation}
where $L_{xe}$ is the cross entropy loss and $Target \in \mathbb{R}^{N\times(E+1)}$, where each element indicates if a region and an entity are matched or not.

\subsection{Implementation Details}
For KeViLI, the input image is resized to $640 \times 640$, and the max (token-based) length for $T$ and $K$ are set to 32 and 256, respectively.
During training, the model is optimized with the batch size set to 64 using AdamW optimizer~\cite{loshchilov2018adamw}, where the initial learning rate of the vision encoder and language encoder is set to $10^{-5}$ and the learning rate of the remaining parameters are set to $10^{-4}$.
Similar to \cite{deng2021transvg}, the vision encoder is initialized from the DETR model \cite{carion2020end}, and the language encoder is initialized with the BERT model \cite{devlin2019bert}.
The model is trained for 90 epochs with a learning rate dropped by a factor of 10 after the 60th epoch.
For LeViLM, we initialize the backbone model from \cite{li2022grounded}.
We train the model with the batch size set to 32.
Similarly, the learning rate is set to $10^{-5}$ for the text encoder and $10^{-4}$ for the remaining parameters.
During training, the learning rate is decayed at 67\% and 89\% of the total training steps.

To evaluate LeViLM on the SK-VG dataset, we testify different experimental settings:
\begin{itemize}[noitemsep,nolistsep]
    \item \textbf{Data}: Query-only (Q), Query-knowledge (Q+K), and Query-knowledge-linguistic-structure (Q+K+S);
    \item \textbf{Training}: (i) Zero-shot (ZS): We directly evaluated the pre-trained model without finetuning; (ii) Linear-probing (LP): We fixed the backbone model; (iii) Finetuning (FT): We tuned all the parameters of the model.
    \item \textbf{Evaluation}\footnote{Since the {LeViLM} model might predict multiple bounding boxes, we adapt different evaluation strategies for analysis.}: (i) Selecting the prediction with the highest score (H), (ii) Randomly picking the prediction whose score is over 0.5 (R), (iii) Selecting the ground-truth one if its score is larger than 0.5 (U).\footnote{The U strategy is adopted to analyze the reasoning error of the model.}
\end{itemize}

For the evaluation metric, we adopt Intersection-over-Union (IoU), which measures the overlap degree between the prediction and the ground truth. Following previous studies \cite{yu2016modeling,mao2016generation,deng2021transvg}, we use IOU@0.5 as the prediction accuracy.

\subsection{Experimental Results and Analyses}
\begin{table}[t]
\centering
\footnotesize
\resizebox{0.99\linewidth}{!}{
\begin{tabular}{@{}lccccccc@{}}
\toprule
Method & \cite{yu2018mattnet} & \cite{liu2019learning} & \cite{liu2019improving} & \cite{yang2019fast} & \cite{yang2020improving} & KeViLI & LeViLM \\ \midrule
Acc    & 25.28                & 25.24                  & 26.08                   & 16.3                & 36.68                    & 30.01  & 72.57  \\ \bottomrule
\end{tabular}}
\caption{Comparisons of our approaches with existing studies.}
\label{table:comparisons}
\end{table}
\begin{table*}[t]
\centering
\footnotesize
\resizebox{0.83\linewidth}{!}{\begin{tabular}{@{}lllccccccccc@{}}
\toprule
\multicolumn{2}{l}{}                                  &                                 &                                     &                               &                                        & \multicolumn{3}{c}{\textbf{Difficulty-level}}                                              & \multicolumn{3}{c}{\textbf{Area-level}}                                                    \\
\multicolumn{2}{l}{\multirow{-2}{*}{\textbf{Method}}} & \multirow{-2}{*}{\textbf{Text}} & \multirow{-2}{*}{\textbf{Criteria}} & \multirow{-2}{*}{\textbf{ID}} & \multirow{-2}{*}{\textbf{Overall Acc}} & \textbf{Acc$_{de}$}          & \textbf{Acc$_{dm}$}          & \textbf{Acc$_{dh}$}          & \textbf{Acc$_{as}$}          & \textbf{Acc$_{am}$}          & \textbf{Acc$_{al}$}          \\ \midrule
\multicolumn{2}{l}{}                                  & Q                               & -                                   & 1                             & 28.71                   & \multicolumn{1}{c}{32.53}         & \multicolumn{1}{c}{25.23}         & \multicolumn{1}{c}{25.70}         & \multicolumn{1}{c}{0.80}         & \multicolumn{1}{c}{14.44}         & 34.02         \\
\multicolumn{2}{l}{\multirow{-2}{*}{{KeViLI}}}         & Q~+~K                           & -                                   & 2                             & 30.01                   & \multicolumn{1}{c}{33.75}         & \multicolumn{1}{c}{26.55}         & \multicolumn{1}{c}{27.14}         & \multicolumn{1}{c}{1.20}         & \multicolumn{1}{c}{12.85}         & 35.94         \\ \midrule
\multicolumn{2}{l}{}                                  &                                 & H                                   & 3                             & 29.75                                  & 49.97                        & 18.23                        & 6.71                         & 24.20                        & 33.33                        & 29.64                        \\
\multicolumn{2}{l}{}                                  &                                 & R                                   & 4                             & 29.77                                  & 48.28                        & 18.88                        & 9.01                         & 23.20                        & 33.12                        & 29.79                        \\
\multicolumn{2}{l}{}                                  & \multirow{-3}{*}{Q}             & {\color[HTML]{9B9B9B} U}            & {\color[HTML]{9B9B9B} 5}      & {\color[HTML]{9B9B9B} 38.13}           & {\color[HTML]{9B9B9B} 54.23} & {\color[HTML]{9B9B9B} 29.56} & {\color[HTML]{9B9B9B} 19.16} & {\color[HTML]{9B9B9B} 30.20} & {\color[HTML]{9B9B9B} 39.38} & {\color[HTML]{9B9B9B} 38.67} \\ \cmidrule(l){3-12} 
\multicolumn{2}{l}{}                                  &                                 & H                                   & 6                             & 7.55                                   & 13.08                        & 4.38                         & 1.26                         & 2.20                         & 5.94                         & 8.36                         \\
\multicolumn{2}{l}{}                                  &                                 & R                                   & 7                             & 7.78                                   & 12.88                        & 4.71                         & 2.12                         & 2.20                         & 5.73                         & 8.69                         \\
\multicolumn{2}{l}{\multirow{-6}{*}{{LeViLM} (ZS)}}       & \multirow{-3}{*}{Q~+~K}         & {\color[HTML]{9B9B9B} U}            & {\color[HTML]{9B9B9B} 8}      & {\color[HTML]{9B9B9B} 8.79}            & {\color[HTML]{9B9B9B} 13.34} & {\color[HTML]{9B9B9B} 6.02}  & {\color[HTML]{9B9B9B} 3.79}  & {\color[HTML]{9B9B9B} 2.20}  & {\color[HTML]{9B9B9B} 6.05}  & {\color[HTML]{9B9B9B} 9.93}  \\ \midrule
\multicolumn{2}{l}{}                                  &                                 & H                                   & 9                             & 44.97                                  & 72.03                        & 31.86                        & 11.70                        & 50.60                        & 57.54                        & 42.13                        \\
\multicolumn{2}{l}{}                                  &                                 & R                                   & 10                            & 44.82                                  & 66.91                        & 32.68                        & 19.16                        & 48.20                        & 56.48                        & 42.36                        \\
\multicolumn{2}{l}{}                                  & \multirow{-3}{*}{Q}             & {\color[HTML]{9B9B9B} U}            & {\color[HTML]{9B9B9B} 11}     & {\color[HTML]{9B9B9B} 63.09}           & {\color[HTML]{9B9B9B} 77.51} & {\color[HTML]{9B9B9B} 54.90} & {\color[HTML]{9B9B9B} 46.64} & {\color[HTML]{9B9B9B} 52.60} & {\color[HTML]{9B9B9B} 64.86} & {\color[HTML]{9B9B9B} 63.79} \\ \cmidrule(l){3-12} 
\multicolumn{2}{l}{}                                  &                                 & H                                   & 12                            & 35.71                                  & 60.40                        & 25.07                        & 3.96                         & 41.20                        & 48.51                        & 32.84                        \\
\multicolumn{2}{l}{}                                  &                                 & R                                   & 13                            & 35.89                                  & 57.00                        & 24.41                        & 11.24                        & 39.40                        & 47.13                        & 33.49                        \\
\multicolumn{2}{l}{}                                  & \multirow{-3}{*}{Q~+~K}         & {\color[HTML]{9B9B9B} U}            & {\color[HTML]{9B9B9B} 14}     & {\color[HTML]{9B9B9B} 47.71}           & {\color[HTML]{9B9B9B} 64.40} & {\color[HTML]{9B9B9B} 41.43} & {\color[HTML]{9B9B9B} 25.30} & {\color[HTML]{9B9B9B} 43.20} & {\color[HTML]{9B9B9B} 55.52} & {\color[HTML]{9B9B9B} 46.72} \\ \cmidrule(l){3-12} 
\multicolumn{2}{l}{}                                  &                                 & H                                   & 15                            & 37.25                                  & 62.09                        & 26.98                        & 4.88                         & 42.20                        & 49.47                        & 34.54                        \\
\multicolumn{2}{l}{}                                  &                                 & R                                   & 16                            & 36.91                                  & 58.03                        & 25.83                        & 11.82                        & 40.20                        & 46.92                        & 34.76                        \\
\multicolumn{2}{l}{\multirow{-9}{*}{{LeViLM} (LP)}}       & \multirow{-3}{*}{Q~+~K~+~S}     & {\color[HTML]{9B9B9B} U}            & {\color[HTML]{9B9B9B} 17}     & {\color[HTML]{9B9B9B} 50.47}           & {\color[HTML]{9B9B9B} 66.61} & {\color[HTML]{9B9B9B} 44.77} & {\color[HTML]{9B9B9B} 28.40} & {\color[HTML]{9B9B9B} 44.40} & {\color[HTML]{9B9B9B} 56.69} & {\color[HTML]{9B9B9B} 49.92} \\ \midrule
\multicolumn{2}{l}{}                                  &                                 & H                                   & 18                            & 57.18                                  & 80.35                        & 46.80                        & 27.83                        & 65.00                        & 66.77                        & 54.67                        \\
\multicolumn{2}{l}{}                                  &                                 & R                                   & 19                            & 57.29                                  & 80.15                        & 46.63                        & 28.74                        & 65.00                        & 65.39                        & 55.06                        \\
\multicolumn{2}{l}{}                                  & \multirow{-3}{*}{Q}             & {\color[HTML]{9B9B9B} U}            & {\color[HTML]{9B9B9B} 20}     & {\color[HTML]{9B9B9B} 63.79}           & {\color[HTML]{9B9B9B} 83.45} & {\color[HTML]{9B9B9B} 55.17} & {\color[HTML]{9B9B9B} 38.67} & {\color[HTML]{9B9B9B} 68.60} & {\color[HTML]{9B9B9B} 71.23} & {\color[HTML]{9B9B9B} 61.97} \\ \cmidrule(l){3-12} 
\multicolumn{2}{l}{}                                  &                                 & H                                   & 21                            & 70.70                                  & 84.51                        & 63.16                        & 54.62                        & 68.20                        & 72.51                        & 70.62                        \\
\multicolumn{2}{l}{}                                  &                                 & R                                   & 22                            & 70.49                                  & 84.28                        & 62.67                        & 54.73                        & 68.80                        & 72.51                        & 70.29                        \\
\multicolumn{2}{l}{}                                  & \multirow{-3}{*}{Q~+~K}         & {\color[HTML]{9B9B9B} U}            & {\color[HTML]{9B9B9B} 23}     & {\color[HTML]{9B9B9B} 74.95}           & {\color[HTML]{9B9B9B} 86.49} & {\color[HTML]{9B9B9B} 68.20} & {\color[HTML]{9B9B9B} 61.96} & {\color[HTML]{9B9B9B} 71.00} & {\color[HTML]{9B9B9B} 76.11} & {\color[HTML]{9B9B9B} 75.12} \\ \cmidrule(l){3-12} 
\multicolumn{2}{l}{}                                  &                                 & H                                   & 24                            & 72.57                                  & 84.08                        & 65.52                        & 59.95                        & 70.00                        & 71.02                        & 73.10                        \\
\multicolumn{2}{l}{}                                  &                                 & R                                   & 25                            & 71.93                                  & 83.72                        & 64.97                        & 58.75                        & 70.00                        & 71.44                        & 72.21                        \\
\multicolumn{2}{l}{\multirow{-9}{*}{{LeViLM} (FT)}}       & \multirow{-3}{*}{Q~+~K~+~S}     & {\color[HTML]{9B9B9B} U}            & {\color[HTML]{9B9B9B} 26}     & {\color[HTML]{9B9B9B} 77.31}           & {\color[HTML]{9B9B9B} 86.59} & {\color[HTML]{9B9B9B} 71.59} & {\color[HTML]{9B9B9B} 67.18} & {\color[HTML]{9B9B9B} 72.60} & {\color[HTML]{9B9B9B} 76.96} & {\color[HTML]{9B9B9B} 77.83} \\ \bottomrule
\end{tabular}}
\caption{The performance of two proposed approaches.
In the text column, Q, K, and S represent query, knowledge, and linguistic structure, respectively.
In the criteria column, H, R, and U represent the criteria to pick the detected bounding boxes by adopting the boxes with the highest scores, the random boxes, and the upper-bound scores that can be achieved, respectively.
For the metrics, the overall accuracy, the difficulty-level accuracy, and the area-level accuracy are shown.}
\label{table:baselines}
\end{table*}
To analyze the performance of different baselines, we consider the following questions and conduct analyses to answer them with the results reported in Table~\ref{table:comparisons} and \ref{table:baselines}.

\noindent\textit{Q1: Is SK-VG a hard task for traditional VG models?}
As shown in Table~\ref{table:comparisons}, existing models did not achieve promising results
An interesting finding is that ReSC, which uses texts to recursively refine the text-conditional visual features, achieved the best result ($\sim$36\%) among these existing models, which matches our intuition that it can better use long-form story information.

\noindent\textit{Q2: Which one is better, KeViLI or LeViLM?}
It can be observed in Table~\ref{table:baselines} that the performance of LeViLM (ID 3-26) is consistently better than KeViLI (ID 1-2), even without any finetuning (ID 3-4).
We can explain this by the reason that the task's inherent difficulty is understanding open-ended stories, queries, and their relations with the images.
For KeViL, the one-stage optimization to directly output bounding boxes of such open-ended target objects could be difficult.
Instead, for LeViLM, after dividing and conquering the process (i.e., region proposing and scoring), it is easier to ensure each stage works well, e.g., to guarantee its basic detection ability before complex grounding using a pre-trained VG backbone.

\noindent\textit{Q3: Are linear-probing or finetuning necessary for LeViLM?}
We can investigate the effects of linear probing and finetuning by comparing the results (ID 3-8, ID 9-14, and ID 18-23).
When adapting LeViLM on this dataset, the performance follows this pattern: finetuning~\textgreater~linear-probing~\textgreater~zero-shot.
The reason behind this is that finetuning can guide the model to use the scene knowledge in a better way.

\noindent\textit{Q4: Is the scene knowledge critical for accurate prediction?}
To answer this question, we need to take the different evaluation strategies into account.
Specifically, in the ZS and LP setting, it can be observed that the knowledge is harmful to the model performance by comparing ID 3-5 and ID 6-8 (or comparing ID 9-11 and ID 12-17).
This is due to two reasons: (i) The texts of the pretraining datasets of LeViLM are relatively short, yet the length of scene knowledge in our dataset is much longer than that; (ii) The majority of the LeViLM pretraining datasets are about perception, i.e., detecting all the objects in the images instead of reasoning over the images and texts.
Therefore, it is not enough to exploit the knowledge under the zero-shot and linear-probing settings.
On the contrary, the knowledge has a considerably positive effect when full-finetuning {LeViLM} on the proposed dataset, which can be explained by the fact that LeViLM learns to reason over the images, knowledge, and querying texts after adaptation.
Besides, bridging the scene knowledge and the queries in an appropriate way (ID 24-25) can further promote performance.
The conclusion is that knowledge is critical for finetuning but can not be exploited appropriately in the zero-shot and linear-probing settings.

% ****************** Figure 2 ******************
\begin{figure*}[t]
\centering
\includegraphics[width=0.9\textwidth, trim=0 10 0 0]{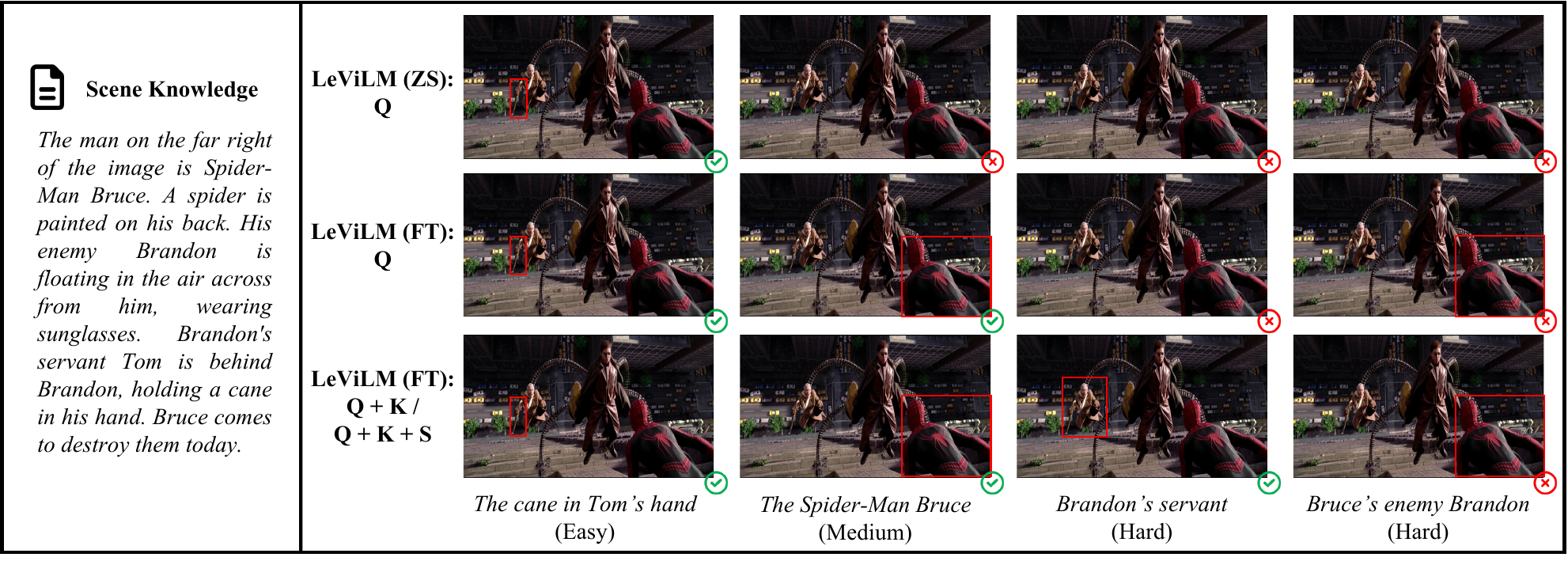}
\caption{The illustration of samples from the proposed SK-VG dataset, where a scene story and its four referring expressions are shown with the grounding results from four baseline methods.}
\label{fig:case_study}
\end{figure*}
% ****************** Figure 2 ******************

\noindent\textit{Q5: What is the advantage of exploiting the knowledge?}
For this question, there are two interesting observations from the results.
First, when using the knowledge, the model can achieve a higher upper-bound result (comparing ID 20, 23, and 26), which means that the model can detect more objects in the images.
We can explain this phenomenon by one of the possible instances: the querying text might contain different names of persons, and the model might not know the name refers to a person, which can be inferred from the knowledge.
Second, with the knowledge, the model is able to perform more accurate reasoning, which can be observed by comparing the reasoning errors (the results of $U - H$) of Q and Q+K (or Q+K+S).
This is because the knowledge can alleviate the reasoning uncertainty when grounding the objects.
The answer is that scene knowledge can not only assist in detecting more objects but also reduce the uncertainty of locating/reasoning the target objects.

\noindent\textit{Q6: What do the approaches still struggle to do?}
Before answering this question, we can investigate the effects of the area of objects.
As shown in Table~\ref{table:baselines}, it is not challenging for LeViLM to detect small objects.
By observing the difficulty-level accuracy, we can obtain the message that {LeViLM} is not capable of performing complicated (multi-hop) reasoning over the scene knowledge and producing accurate predictions.
Besides, the prediction process is black-box and can not be explainable, which can be further studied in the future.
The answer is that (i) The current baselines can only achieve strong results on easy or medium tasks and are unable to perform well on the hard task; (ii) The interpretability of the baselines is poor.

\subsection{Case Study}
\label{subsec:case_study}
To further investigate the effects of knowledge, we perform qualitative analysis on four cases in the SK-VG dataset.
Figure \ref{fig:case_study} shows the grounding results of four baselines on four referring expressions.
It is observed that in the first case, all the baselines can ground the ``\textit{cane}'' in the image even without the knowledge since there is only one cane presented.
In the second case, the finetuned {LeViLM} can detect the target object even without knowledge, while it can not detect the ``\textit{Brandon's servant}'' without knowledge in the third case.
In the last case, all the baselines can not ground the referred object correctly, and the last three baselines all treat the ``\textit{Spider-Man}'' as the ``\textit{enemy}''.
This shows that the baseline models can not perform accurate reasoning in some complicated cases, demonstrating the challenges.

\section{Concluding Remarks}
\label{sec:conclusion}
The visual grounding field has emerged as a prominent attractive research direction, where the models are required to reason over vision and language to ground the target objects.
Yet, the language part of the existing VG benchmarks is only simple description texts, which can not evaluate the reasoning capability of the models comprehensively.
To take a step in this direction, we propose a new benchmark dataset called SK-VG, which requires models to reason over the (image, scene knowledge, query) triples to perform accurate reasoning.
We propose two approaches to perform this new task: Knowledge-embedded Vision-Language Interaction and Linguistic-enhanced Vision-Language Matching.
Experimental results confirm the validity of the proposed approaches but also show that there is still substantial room for improvement, e.g., reasoning and interpretability.

\section*{Acknowledgement}
This work was supported in part by the Chinese Key-Area Research and Development Program of Guangdong Province (2020B0101350001), in part by the Guangdong Basic and Applied Basic Research Foundation (NO.~2020B1515020048), in part by the National Natural Science Foundation of China (NO.~61976250), in part by the Shenzhen Science and Technology Program (NO.~JCYJ20220530141211024, NO.~JCYJ20220818103001002), in part by the Fundamental Research Funds for the Central Universities under Grant 22lgqb25 and in part by the Guangdong Provincial Key Laboratory of Big Data Computing, The Chinese University of Hong Kong, Shenzhen. This work was also sponsored by Tencent CCF Open Fund (NO.~RBFR2022009).

\clearpage

{\small
\bibliographystyle{ieee_fullname}
\bibliography{cvpr}}

\begin{thebibliography}{10}\itemsep=-1pt

\bibitem{anderson2018vision}
Peter Anderson, Qi Wu, Damien Teney, Jake Bruce, Mark Johnson, Niko
  S{\"u}nderhauf, Ian Reid, Stephen Gould, and Anton Van Den~Hengel.
\newblock Vision-and-language navigation: Interpreting visually-grounded
  navigation instructions in real environments.
\newblock In {\em Proceedings of the IEEE conference on computer vision and
  pattern recognition}, pages 3674--3683, 2018.

\bibitem{bajaj2019g3raphground}
Mohit Bajaj, Lanjun Wang, and Leonid Sigal.
\newblock G3raphground: Graph-based language grounding.
\newblock In {\em Proceedings of the IEEE/CVF International Conference on
  Computer Vision}, pages 4281--4290, 2019.

\bibitem{carion2020end}
Nicolas Carion, Francisco Massa, Gabriel Synnaeve, Nicolas Usunier, Alexander
  Kirillov, and Sergey Zagoruyko.
\newblock End-to-end object detection with transformers.
\newblock In {\em European conference on computer vision}, pages 213--229.
  Springer, 2020.

\bibitem{chen2021grounding}
Zhenfang Chen, Jiayuan Mao, Jiajun Wu, Kwan-Yee~Kenneth Wong, Joshua~B
  Tenenbaum, and Chuang Gan.
\newblock Grounding physical concepts of objects and events through dynamic
  visual reasoning.
\newblock In {\em ICLR}, 2021.

\bibitem{chen2020cops}
Zhenfang Chen, Peng Wang, Lin Ma, Kwan-Yee~K Wong, and Qi Wu.
\newblock Cops-ref: A new dataset and task on compositional referring
  expression comprehension.
\newblock In {\em Proceedings of the IEEE/CVF Conference on Computer Vision and
  Pattern Recognition}, pages 10086--10095, 2020.

\bibitem{chencomphy}
Zhenfang Chen, Kexin Yi, Yunzhu Li, Mingyu Ding, Antonio Torralba, Joshua~B
  Tenenbaum, and Chuang Gan.
\newblock Comphy: Compositional physical reasoning of objects and events from
  videos.
\newblock In {\em ICLR}, 2022.

\bibitem{dai2021dynamic}
Xiyang Dai, Yinpeng Chen, Bin Xiao, Dongdong Chen, Mengchen Liu, Lu Yuan, and
  Lei Zhang.
\newblock Dynamic head: Unifying object detection heads with attentions.
\newblock In {\em Proceedings of the IEEE/CVF conference on computer vision and
  pattern recognition}, pages 7373--7382, 2021.

\bibitem{deng2021transvg}
Jiajun Deng, Zhengyuan Yang, Tianlang Chen, Wengang Zhou, and Houqiang Li.
\newblock Transvg: End-to-end visual grounding with transformers.
\newblock In {\em Proceedings of the IEEE/CVF International Conference on
  Computer Vision}, pages 1769--1779, 2021.

\bibitem{devlin2019bert}
Jacob Devlin, Ming-Wei Chang, Kenton Lee, and Kristina Toutanova.
\newblock Bert: Pre-training of deep bidirectional transformers for language
  understanding.
\newblock In {\em Proceedings of the 2019 Conference of the North American
  Chapter of the Association for Computational Linguistics: Human Language
  Technologies, Volume 1 (Long and Short Papers)}, pages 4171--4186, 2019.

\bibitem{ding2021dynamic}
Mingyu Ding, Zhenfang Chen, Tao Du, Ping Luo, Josh Tenenbaum, and Chuang Gan.
\newblock Dynamic visual reasoning by learning differentiable physics models
  from video and language.
\newblock {\em Advances In Neural Information Processing Systems}, 34:887--899,
  2021.

\bibitem{ding2022embodied}
Mingyu Ding, Yan Xu, Zhenfang Chen, David~Daniel Cox, Ping Luo, Joshua~B.
  Tenenbaum, and Chuang Gan.
\newblock Embodied concept learner: Self-supervised learning of concepts and
  mapping through instruction following.
\newblock In {\em 6th Annual Conference on Robot Learning}, 2022.

\bibitem{everingham2010pascal}
Mark Everingham, Luc Van~Gool, Christopher~KI Williams, John Winn, and Andrew
  Zisserman.
\newblock The pascal visual object classes (voc) challenge.
\newblock {\em International journal of computer vision}, 88(2):303--338, 2010.

\bibitem{gan2017vqs}
Chuang Gan, Yandong Li, Haoxiang Li, Chen Sun, and Boqing Gong.
\newblock Vqs: Linking segmentations to questions and answers for supervised
  attention in vqa and question-focused semantic segmentation.
\newblock In {\em Proceedings of the IEEE international conference on computer
  vision}, pages 1811--1820, 2017.

\bibitem{gupta2019lvis}
Agrim Gupta, Piotr Dollar, and Ross Girshick.
\newblock Lvis: A dataset for large vocabulary instance segmentation.
\newblock In {\em Proceedings of the IEEE/CVF conference on computer vision and
  pattern recognition}, pages 5356--5364, 2019.

\bibitem{huang2021look}
Binbin Huang, Dongze Lian, Weixin Luo, and Shenghua Gao.
\newblock Look before you leap: Learning landmark features for one-stage visual
  grounding.
\newblock In {\em Proceedings of the IEEE/CVF Conference on Computer Vision and
  Pattern Recognition}, pages 16888--16897, 2021.

\bibitem{kamath2021mdetr}
Aishwarya Kamath, Mannat Singh, Yann LeCun, Gabriel Synnaeve, Ishan Misra, and
  Nicolas Carion.
\newblock Mdetr-modulated detection for end-to-end multi-modal understanding.
\newblock In {\em Proceedings of the IEEE/CVF International Conference on
  Computer Vision}, pages 1780--1790, 2021.

\bibitem{kazemzadeh2014referitgame}
Sahar Kazemzadeh, Vicente Ordonez, Mark Matten, and Tamara Berg.
\newblock Referitgame: Referring to objects in photographs of natural scenes.
\newblock In {\em Proceedings of the 2014 conference on empirical methods in
  natural language processing (EMNLP)}, pages 787--798, 2014.

\bibitem{krishna2017vg}
Ranjay Krishna, Yuke Zhu, Oliver Groth, Justin Johnson, Kenji Hata, Joshua
  Kravitz, Stephanie Chen, Yannis Kalantidis, Li-Jia Li, David~A Shamma, et~al.
\newblock Visual genome: Connecting language and vision using crowdsourced
  dense image annotations.
\newblock {\em International journal of computer vision}, 123(1):32--73, 2017.

\bibitem{li2022grounded}
Liunian~Harold Li, Pengchuan Zhang, Haotian Zhang, Jianwei Yang, Chunyuan Li,
  Yiwu Zhong, Lijuan Wang, Lu Yuan, Lei Zhang, Jenq-Neng Hwang, et~al.
\newblock Grounded language-image pre-training.
\newblock In {\em Proceedings of the IEEE/CVF Conference on Computer Vision and
  Pattern Recognition}, pages 10965--10975, 2022.

\bibitem{lin2014mscoco}
Tsung-Yi Lin, Michael Maire, Serge Belongie, James Hays, Pietro Perona, Deva
  Ramanan, Piotr Doll{\'a}r, and C~Lawrence Zitnick.
\newblock Microsoft coco: Common objects in context.
\newblock In {\em European conference on computer vision}, pages 740--755.
  Springer, 2014.

\bibitem{liu2019learning}
Daqing Liu, Hanwang Zhang, Feng Wu, and Zheng-Jun Zha.
\newblock Learning to assemble neural module tree networks for visual
  grounding.
\newblock In {\em Proceedings of the IEEE/CVF International Conference on
  Computer Vision}, pages 4673--4682, 2019.

\bibitem{liu2020deep}
Li Liu, Wanli Ouyang, Xiaogang Wang, Paul Fieguth, Jie Chen, Xinwang Liu, and
  Matti Pietik{\"a}inen.
\newblock Deep learning for generic object detection: A survey.
\newblock {\em International journal of computer vision}, 128(2):261--318,
  2020.

\bibitem{liu2019clevr}
Runtao Liu, Chenxi Liu, Yutong Bai, and Alan~L Yuille.
\newblock Clevr-ref+: Diagnosing visual reasoning with referring expressions.
\newblock In {\em Proceedings of the IEEE/CVF Conference on Computer Vision and
  Pattern Recognition}, pages 4185--4194, 2019.

\bibitem{liu2016ssd}
Wei Liu, Dragomir Anguelov, Dumitru Erhan, Christian Szegedy, Scott Reed,
  Cheng-Yang Fu, and Alexander~C Berg.
\newblock Ssd: Single shot multibox detector.
\newblock In {\em European conference on computer vision}, pages 21--37.
  Springer, 2016.

\bibitem{liu2019improving}
Xihui Liu, Zihao Wang, Jing Shao, Xiaogang Wang, and Hongsheng Li.
\newblock Improving referring expression grounding with cross-modal
  attention-guided erasing.
\newblock In {\em Proceedings of the IEEE/CVF conference on computer vision and
  pattern recognition}, pages 1950--1959, 2019.

\bibitem{loshchilov2018adamw}
Ilya Loshchilov and Frank Hutter.
\newblock Decoupled weight decay regularization.
\newblock In {\em International Conference on Learning Representations}, 2018.

\bibitem{luo2022goes}
Gen Luo, Yiyi Zhou, Jiamu Sun, Shubin Huang, Xiaoshuai Sun, Qixiang Ye,
  Yongjian Wu, and Rongrong Ji.
\newblock What goes beyond multi-modal fusion in one-stage referring expression
  comprehension: An empirical study.
\newblock {\em arXiv preprint arXiv:2204.07913}, 2022.

\bibitem{luo2020multi}
Gen Luo, Yiyi Zhou, Xiaoshuai Sun, Liujuan Cao, Chenglin Wu, Cheng Deng, and
  Rongrong Ji.
\newblock Multi-task collaborative network for joint referring expression
  comprehension and segmentation.
\newblock In {\em Proceedings of the IEEE/CVF Conference on computer vision and
  pattern recognition}, pages 10034--10043, 2020.

\bibitem{mao2016generation}
Junhua Mao, Jonathan Huang, Alexander Toshev, Oana Camburu, Alan~L Yuille, and
  Kevin Murphy.
\newblock Generation and comprehension of unambiguous object descriptions.
\newblock In {\em Proceedings of the IEEE conference on computer vision and
  pattern recognition}, pages 11--20, 2016.

\bibitem{plummer2015flickr30k}
Bryan~A Plummer, Liwei Wang, Chris~M Cervantes, Juan~C Caicedo, Julia
  Hockenmaier, and Svetlana Lazebnik.
\newblock Flickr30k entities: Collecting region-to-phrase correspondences for
  richer image-to-sentence models.
\newblock In {\em Proceedings of the IEEE international conference on computer
  vision}, pages 2641--2649, 2015.

\bibitem{redmon2016you}
Joseph Redmon, Santosh Divvala, Ross Girshick, and Ali Farhadi.
\newblock You only look once: Unified, real-time object detection.
\newblock In {\em Proceedings of the IEEE conference on computer vision and
  pattern recognition}, pages 779--788, 2016.

\bibitem{ren2015faster}
Shaoqing Ren, Kaiming He, Ross Girshick, and Jian Sun.
\newblock Faster r-cnn: Towards real-time object detection with region proposal
  networks.
\newblock {\em Advances in neural information processing systems}, 28, 2015.

\bibitem{rezatofighi2019generalized}
Hamid Rezatofighi, Nathan Tsoi, JunYoung Gwak, Amir Sadeghian, Ian Reid, and
  Silvio Savarese.
\newblock Generalized intersection over union: A metric and a loss for bounding
  box regression.
\newblock In {\em Proceedings of the IEEE/CVF conference on computer vision and
  pattern recognition}, pages 658--666, 2019.

\bibitem{shao2019objects365}
Shuai Shao, Zeming Li, Tianyuan Zhang, Chao Peng, Gang Yu, Xiangyu Zhang, Jing
  Li, and Jian Sun.
\newblock Objects365: A large-scale, high-quality dataset for object detection.
\newblock In {\em Proceedings of the IEEE/CVF international conference on
  computer vision}, pages 8430--8439, 2019.

\bibitem{vogel2010learning}
Adam Vogel and Dan Jurafsky.
\newblock Learning to follow navigational directions.
\newblock In {\em Proceedings of the 48th annual meeting of the association for
  computational linguistics}, pages 806--814, 2010.

\bibitem{wang2020give}
Peng Wang, Dongyang Liu, Hui Li, and Qi Wu.
\newblock Give me something to eat: referring expression comprehension with
  commonsense knowledge.
\newblock In {\em Proceedings of the 28th ACM International Conference on
  Multimedia}, pages 28--36, 2020.

\bibitem{wang2019neighbourhood}
Peng Wang, Qi Wu, Jiewei Cao, Chunhua Shen, Lianli Gao, and Anton van~den
  Hengel.
\newblock Neighbourhood watch: Referring expression comprehension via
  language-guided graph attention networks.
\newblock In {\em Proceedings of the IEEE/CVF Conference on Computer Vision and
  Pattern Recognition}, pages 1960--1968, 2019.

\bibitem{wang2020general}
Xinyu Wang, Yuliang Liu, Chunhua Shen, Chun~Chet Ng, Canjie Luo, Lianwen Jin,
  Chee~Seng Chan, Anton van~den Hengel, and Liangwei Wang.
\newblock On the general value of evidence, and bilingual scene-text visual
  question answering.
\newblock In {\em Proceedings of the IEEE/CVF Conference on Computer Vision and
  Pattern Recognition}, pages 10126--10135, 2020.

\bibitem{wu2020phrasecut}
Chenyun Wu, Zhe Lin, Scott Cohen, Trung Bui, and Subhransu Maji.
\newblock Phrasecut: Language-based image segmentation in the wild.
\newblock In {\em Proceedings of the IEEE/CVF Conference on Computer Vision and
  Pattern Recognition}, pages 10216--10225, 2020.

\bibitem{yang2019dynamic}
Sibei Yang, Guanbin Li, and Yizhou Yu.
\newblock Dynamic graph attention for referring expression comprehension.
\newblock In {\em Proceedings of the IEEE/CVF International Conference on
  Computer Vision}, pages 4644--4653, 2019.

\bibitem{yang2020graph}
Sibei Yang, Guanbin Li, and Yizhou Yu.
\newblock Graph-structured referring expression reasoning in the wild.
\newblock In {\em Proceedings of the IEEE/CVF Conference on Computer Vision and
  Pattern Recognition}, pages 9952--9961, 2020.

\bibitem{yang2020improving}
Zhengyuan Yang, Tianlang Chen, Liwei Wang, and Jiebo Luo.
\newblock Improving one-stage visual grounding by recursive sub-query
  construction.
\newblock In {\em European Conference on Computer Vision}, pages 387--404.
  Springer, 2020.

\bibitem{yang2022unitab}
Zhengyuan Yang, Zhe Gan, Jianfeng Wang, Xiaowei Hu, Faisal Ahmed, Zicheng Liu,
  Yumao Lu, and Lijuan Wang.
\newblock Unitab: Unifying text and box outputs for grounded vision-language
  modeling.
\newblock In {\em European Conference on Computer Vision}, pages 521--539.
  Springer, 2022.

\bibitem{yang2019fast}
Zhengyuan Yang, Boqing Gong, Liwei Wang, Wenbing Huang, Dong Yu, and Jiebo Luo.
\newblock A fast and accurate one-stage approach to visual grounding.
\newblock In {\em Proceedings of the IEEE/CVF International Conference on
  Computer Vision}, pages 4683--4693, 2019.

\bibitem{you2022humancog}
Haoxuan You, Rui Sun, Zhecan Wang, Kai-Wei Chang, and Shih-Fu Chang.
\newblock Find someone who: Visual commonsense understanding in human-centric
  grounding.
\newblock {\em arXiv preprint arXiv:2212.06971}, 2022.

\bibitem{yu2018mattnet}
Licheng Yu, Zhe Lin, Xiaohui Shen, Jimei Yang, Xin Lu, Mohit Bansal, and
  Tamara~L Berg.
\newblock Mattnet: Modular attention network for referring expression
  comprehension.
\newblock In {\em Proceedings of the IEEE Conference on Computer Vision and
  Pattern Recognition}, pages 1307--1315, 2018.

\bibitem{yu2016modeling}
Licheng Yu, Patrick Poirson, Shan Yang, Alexander~C Berg, and Tamara~L Berg.
\newblock Modeling context in referring expressions.
\newblock In {\em European Conference on Computer Vision}, pages 69--85.
  Springer, 2016.

\bibitem{zellers2019recognition}
Rowan Zellers, Yonatan Bisk, Ali Farhadi, and Yejin Choi.
\newblock From recognition to cognition: Visual commonsense reasoning.
\newblock In {\em Proceedings of the IEEE/CVF conference on computer vision and
  pattern recognition}, pages 6720--6731, 2019.

\bibitem{zhang2022glipv2}
Haotian Zhang, Pengchuan Zhang, Xiaowei Hu, Yen-Chun Chen, Liunian~Harold Li,
  Xiyang Dai, Lijuan Wang, Lu Yuan, Jenq-Neng Hwang, and Jianfeng Gao.
\newblock Glipv2: Unifying localization and vision-language understanding.
\newblock {\em arXiv preprint arXiv:2206.05836}, 2022.

\bibitem{zhou2021real}
Yiyi Zhou, Rongrong Ji, Gen Luo, Xiaoshuai Sun, Jinsong Su, Xinghao Ding,
  Chia-Wen Lin, and Qi Tian.
\newblock A real-time global inference network for one-stage referring
  expression comprehension.
\newblock {\em IEEE Transactions on Neural Networks and Learning Systems},
  2021.

\bibitem{zhu2016visual7w}
Yuke Zhu, Oliver Groth, Michael Bernstein, and Li Fei-Fei.
\newblock Visual7w: Grounded question answering in images.
\newblock In {\em Proceedings of the IEEE conference on computer vision and
  pattern recognition}, pages 4995--5004, 2016.

\end{thebibliography}

\clearpage

\appendix
\appendix
\label{sec:appendix}

\section{Limitations and Future Work}
In this section, we highlight several limitations of this work.
First, the annotation of scene knowledge/stories is a creative process that requires the imagination of the annotators.
Therefore, the scene knowledge annotated by different annotators could vary significantly.
Besides, it could be biased due to the different genders, cultural backgrounds, and interests of the annotators.
Second, compared to existing visual grounding datasets that only require the annotations of the referring expressions, the whole annotation process of the proposed dataset is much more time-consuming.
Although we developed software to improve the annotation efficiency, the scale of our dataset is still smaller than the popular datasets, e.g., RefCOCO, RefCOCO+, and RefCOCOg.
Third, for the methodology, the two-stage approach LeViLM can achieve relatively decent overall performance, especially in the easy/medium split.
However, it still struggles to obtain promising results in the hard split.
%
% Besides, the approach is limited to its poor interpretability of the prediction process.

This work could support and advance the research in the field of visual grounding.
In the future, we can study more techniques (e.g., knowledge distillation) to leverage visual knowledge in the linguistic form to promote the performance of various downstream vision-and-language tasks by embedding knowledge into the visual representations.

\section{The annotation software}
\begin{figure*}[t]
\centering
\includegraphics[width=0.95\textwidth, trim=0 0 0 0]{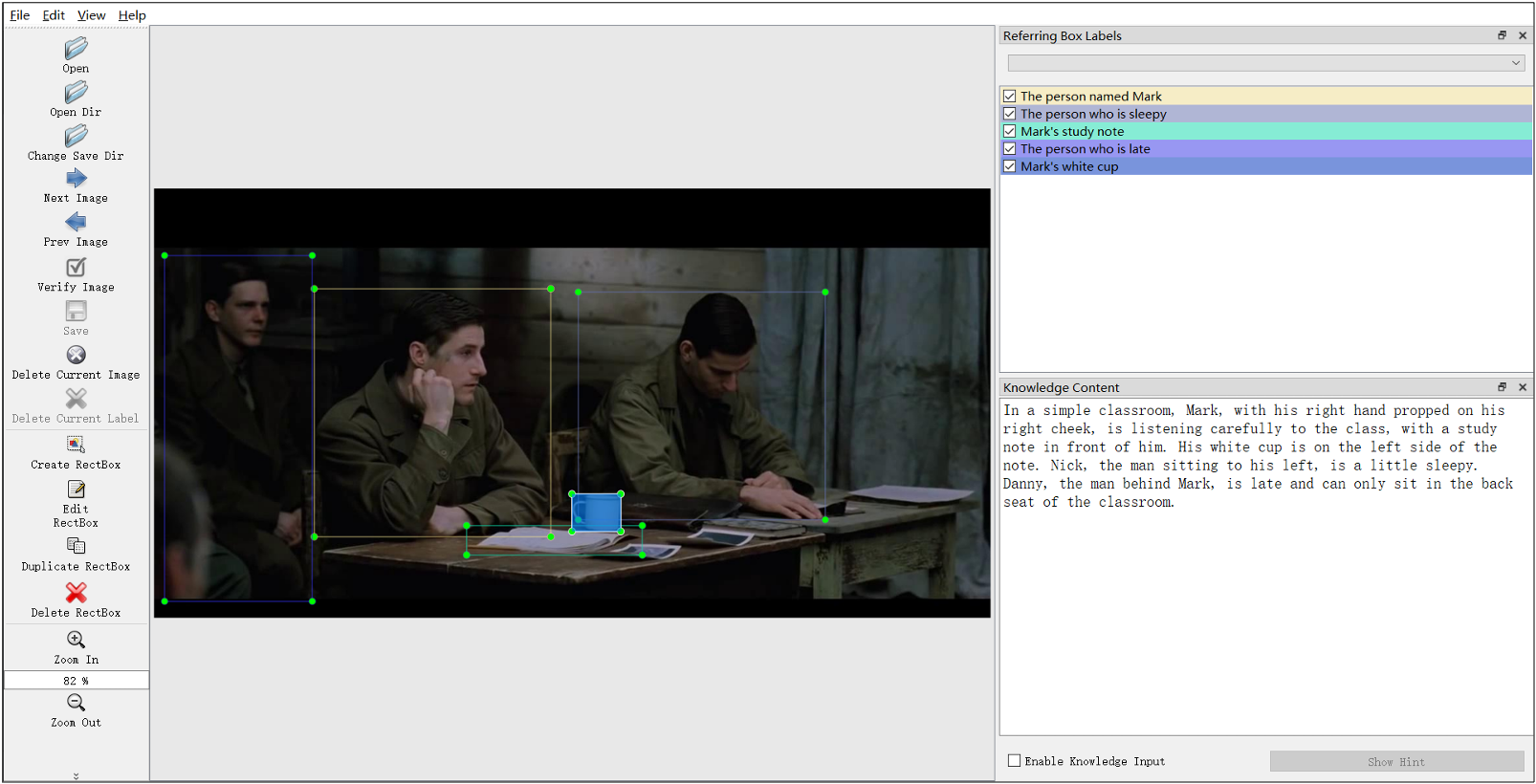}
\caption{The illustration of the annotation software.}
\label{fig:software}
\end{figure*}

We developed software to promote the annotation procedure.
The illustration of the software is shown in Figure~\ref{fig:software}.

\section{More examples}
\begin{figure*}[t]
\centering
\includegraphics[width=0.95\textwidth, trim=0 0 0 0]{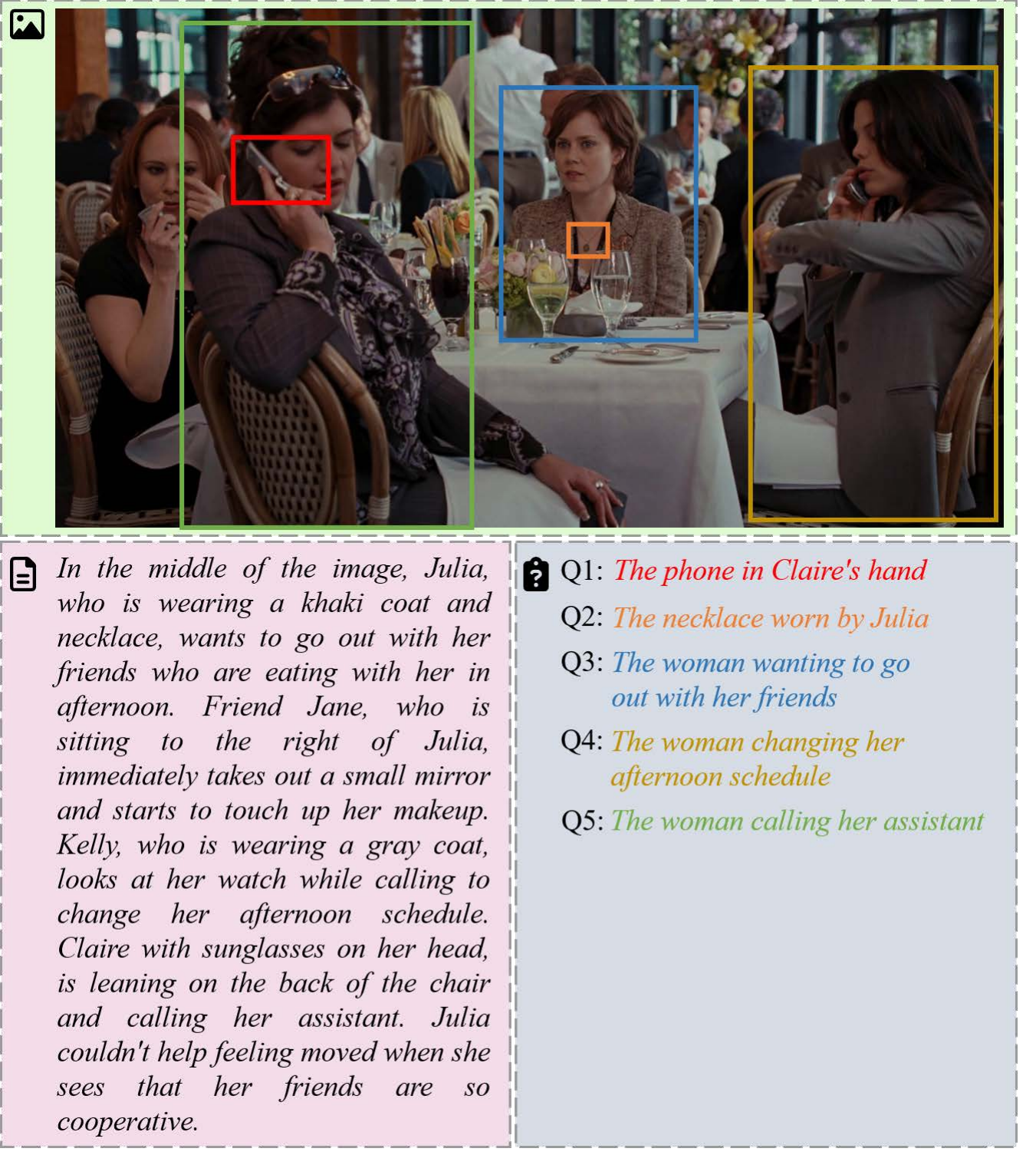}
\caption{An example of the training set.}
\label{fig:example2}
\end{figure*}

\begin{figure*}[t]
\centering
\includegraphics[width=0.95\textwidth, trim=0 0 0 0]{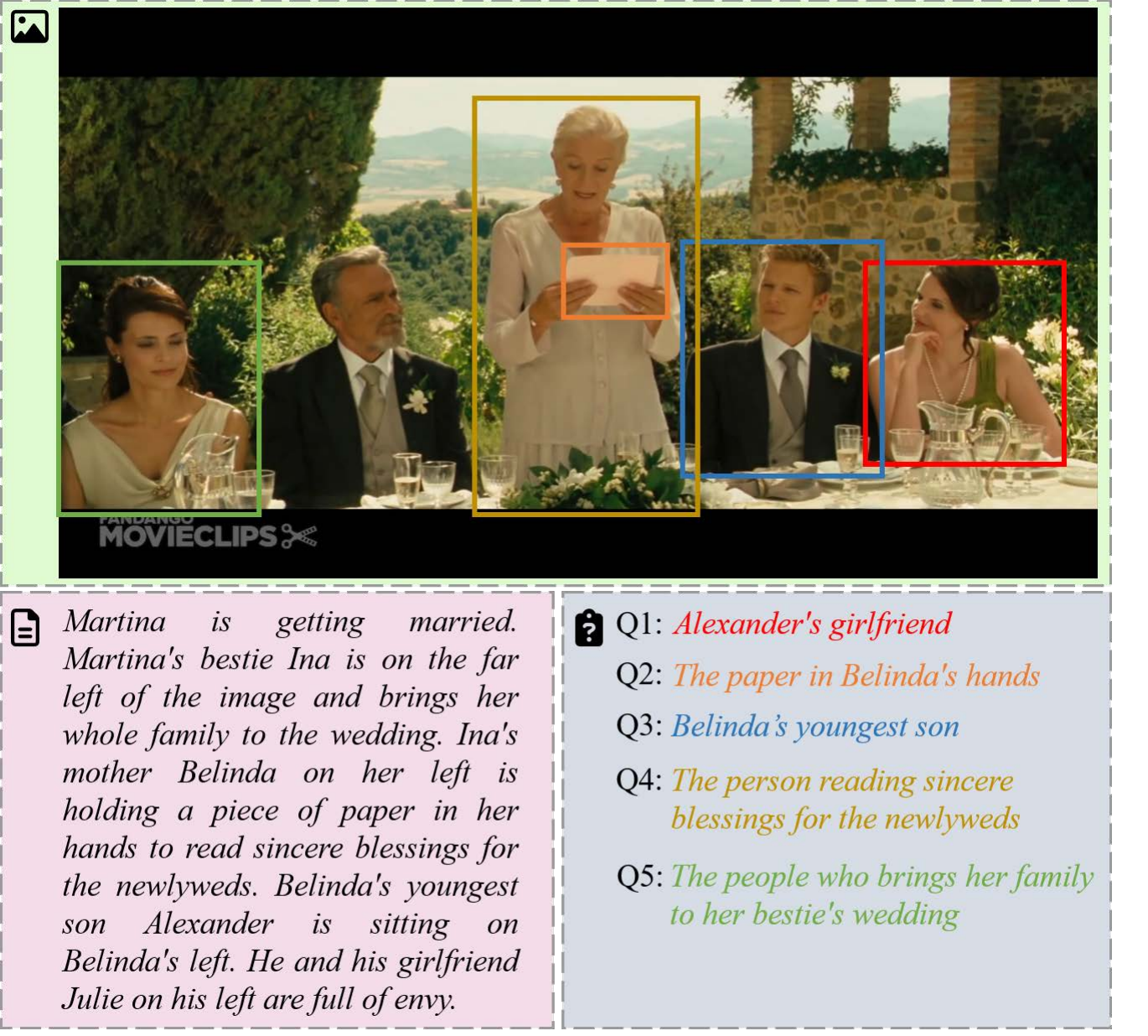}
\caption{An example of the training set.}
\label{fig:example3}
\end{figure*}

\begin{figure*}[t]
\centering
\includegraphics[width=0.95\textwidth, trim=0 0 0 0]{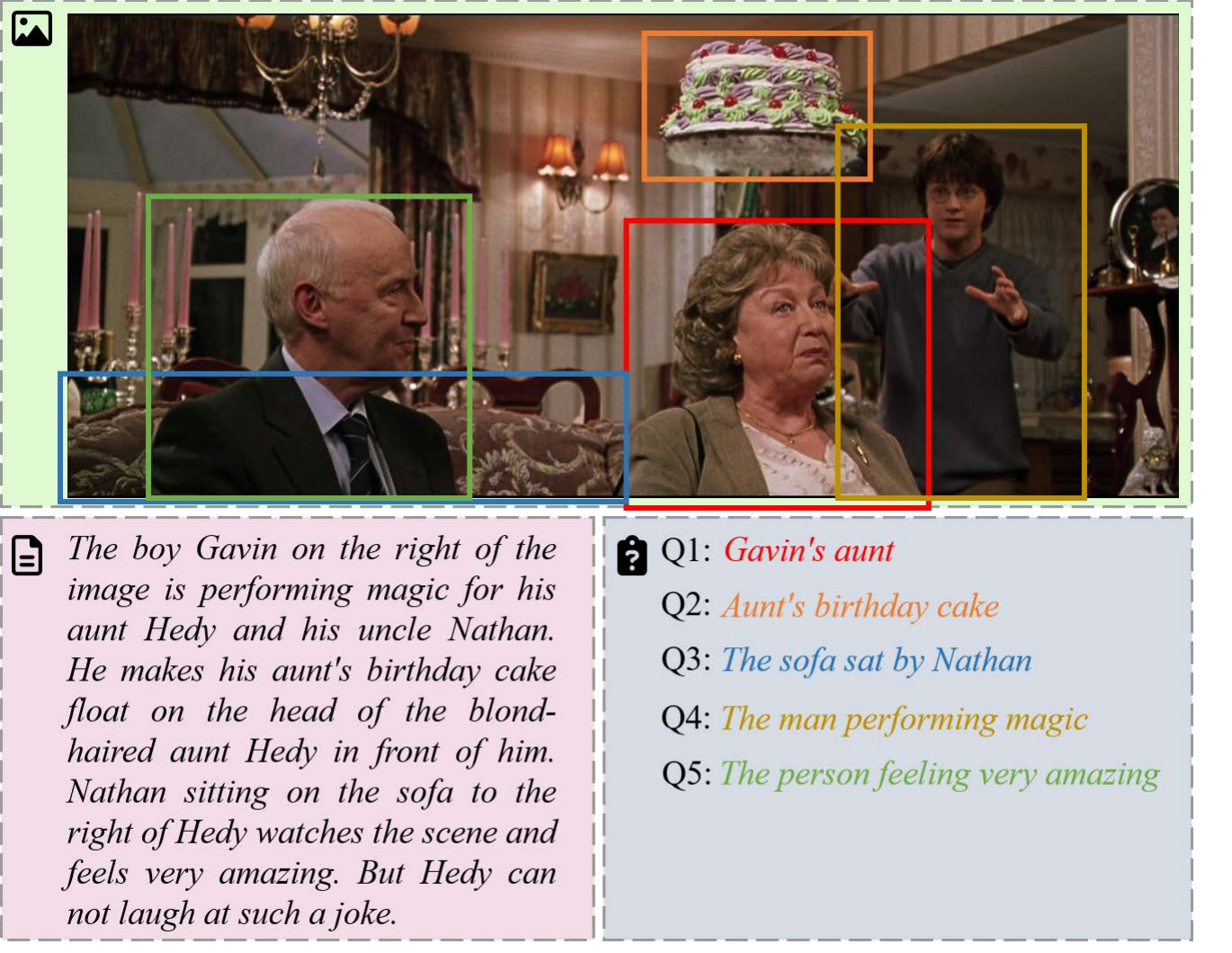}
\caption{An example of the training set.}
\label{fig:example4}
\end{figure*}

\begin{figure*}[t]
\centering
\includegraphics[width=0.95\textwidth, trim=0 0 0 0]{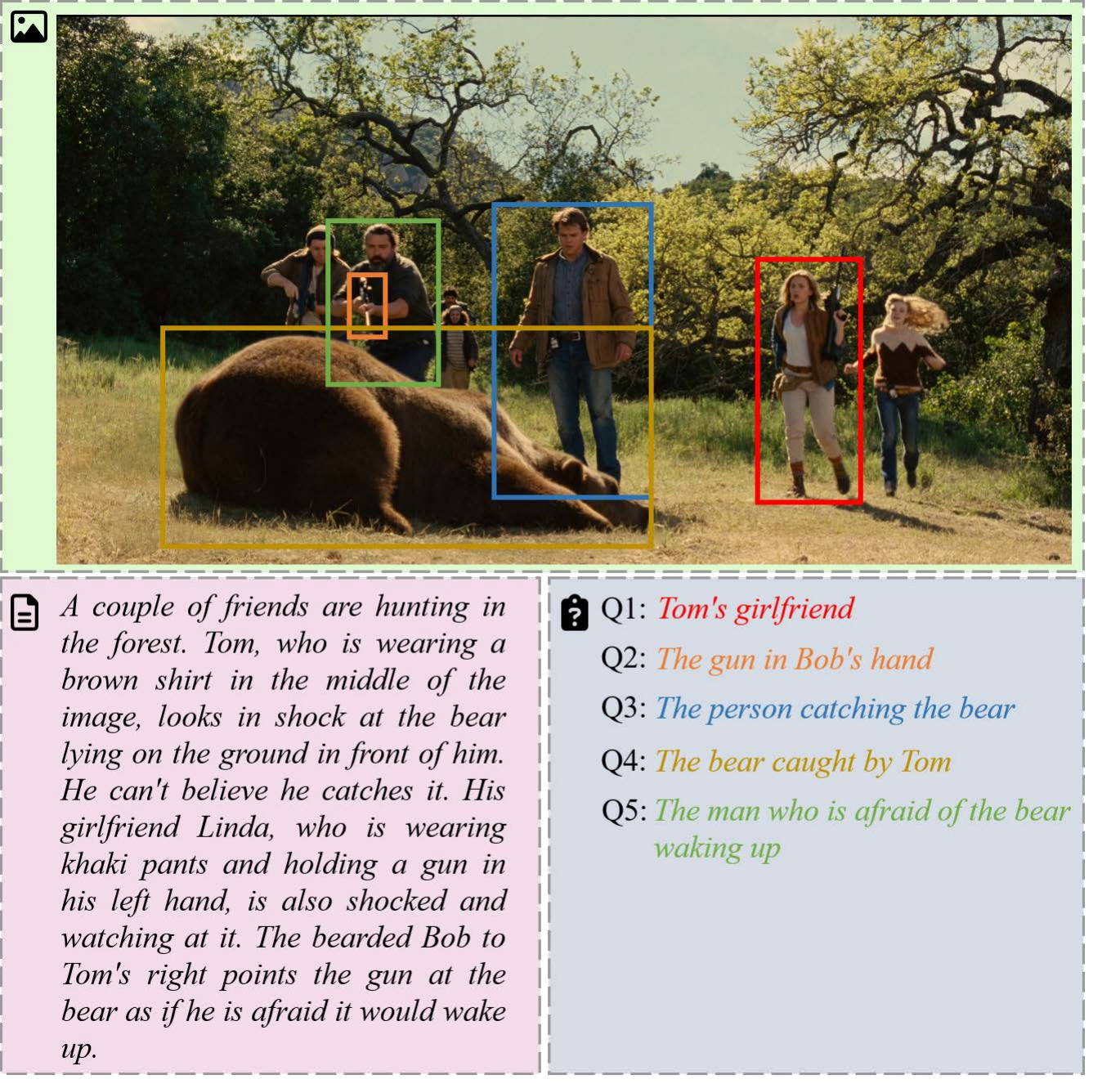}
\caption{An example of the training set.}
\label{fig:example5}
\end{figure*}

\begin{figure*}[t]
\centering
\includegraphics[width=0.95\textwidth, trim=0 0 0 0]{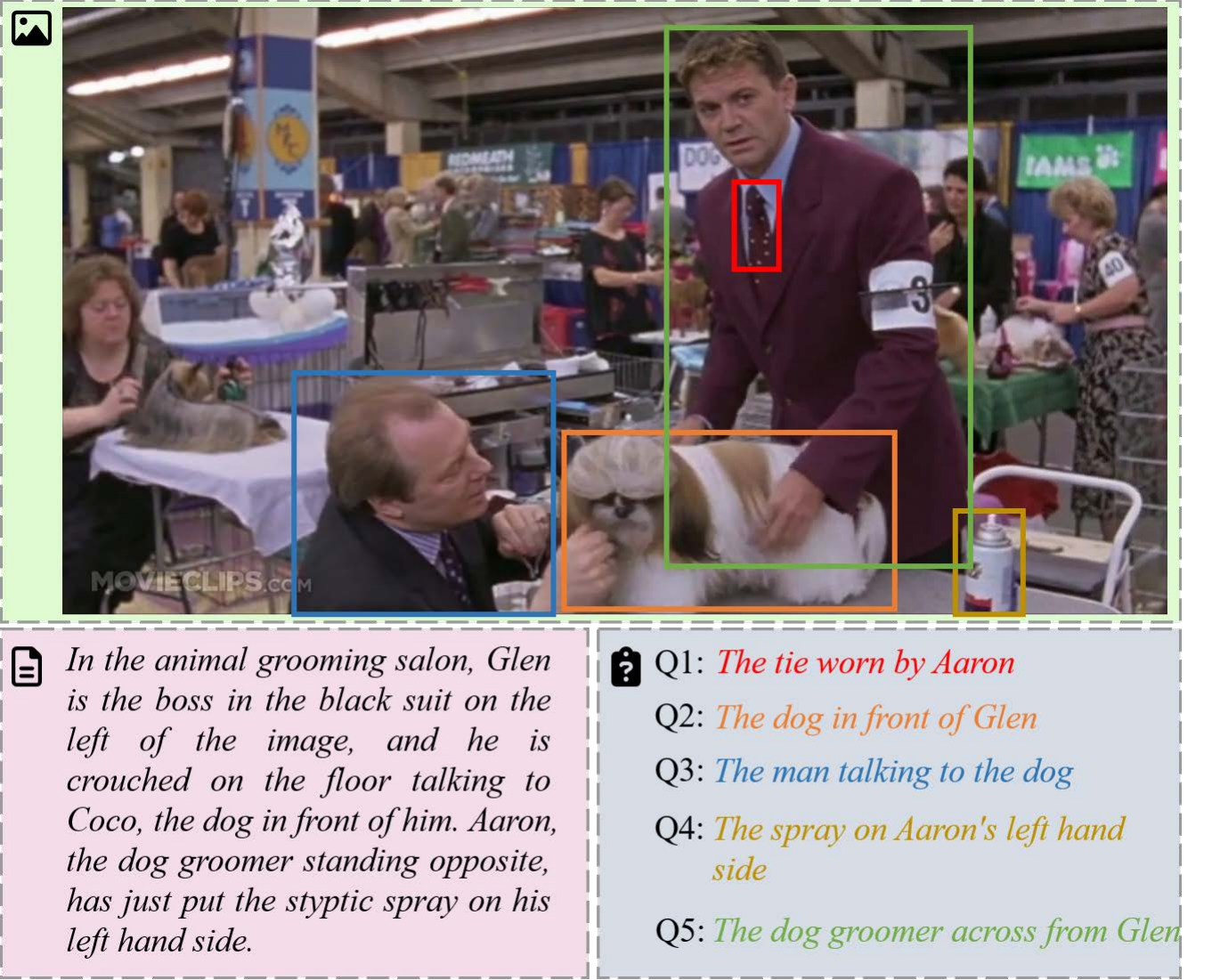}
\caption{An example of the training set.}
\label{fig:example6}
\end{figure*}

\begin{figure*}[t]
\centering
\includegraphics[width=0.95\textwidth, trim=0 0 0 0]{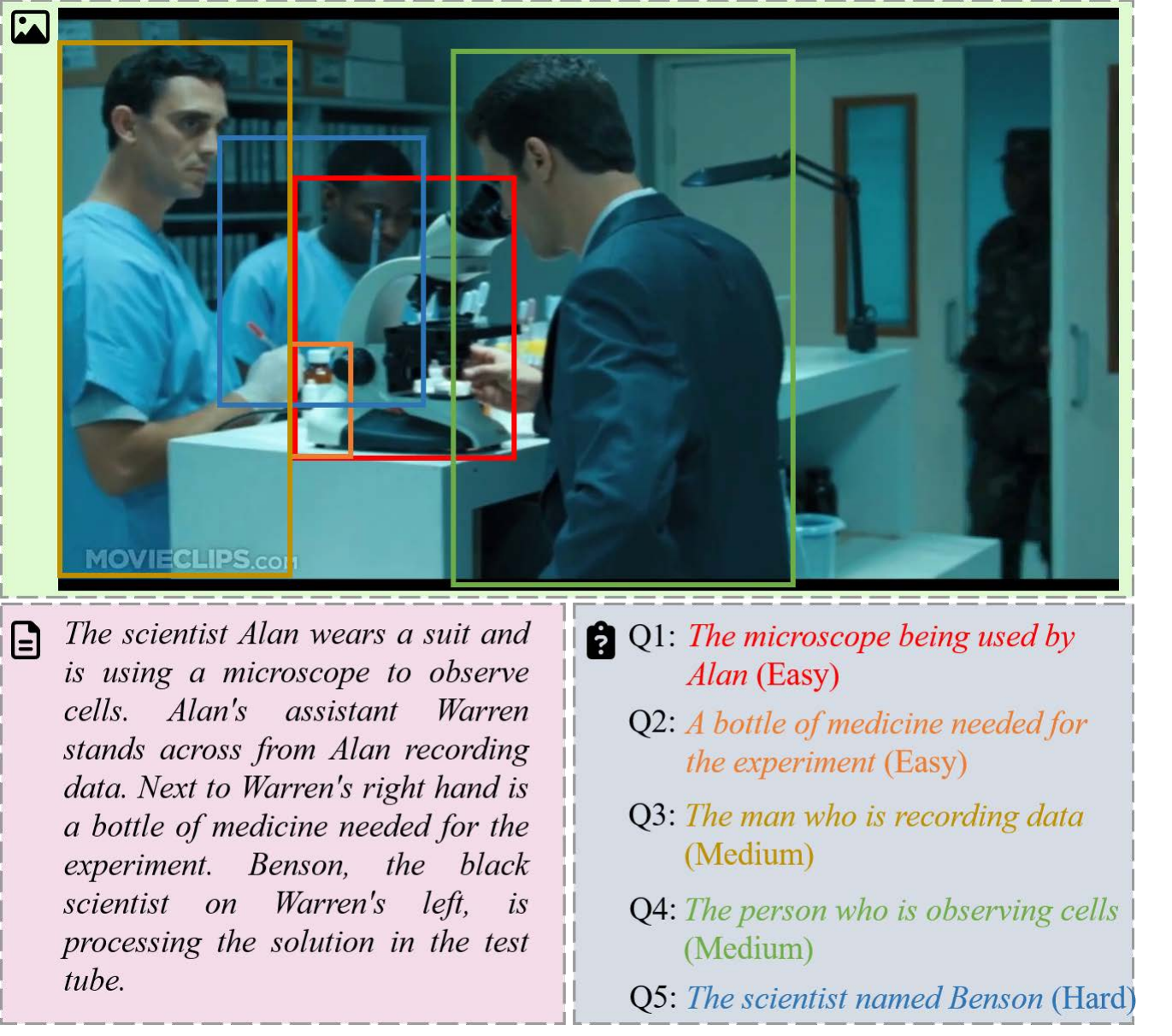}
\caption{An example of the test set. Each query is assigned a difficulty level according to the significance of knowledge for the query. Specifically, Q1 (``\textit{The microscope}'') and Q2 (``\textit{A bottle}'') are visually-obvious and thus classified into \textbf{Easy} level. Q3 and Q4 are knowledge-relevant but provide some visual clues (``\textit{recording data}'' and ``\textit{observing cells}''), rated as \textbf{Medium} level. Q5 relies heavily on the knowledge and deserves a \textbf{Hard} label.}
\label{fig:example7}
\end{figure*}

\begin{figure*}[t]
\centering
\includegraphics[width=0.95\textwidth, trim=0 0 0 0]{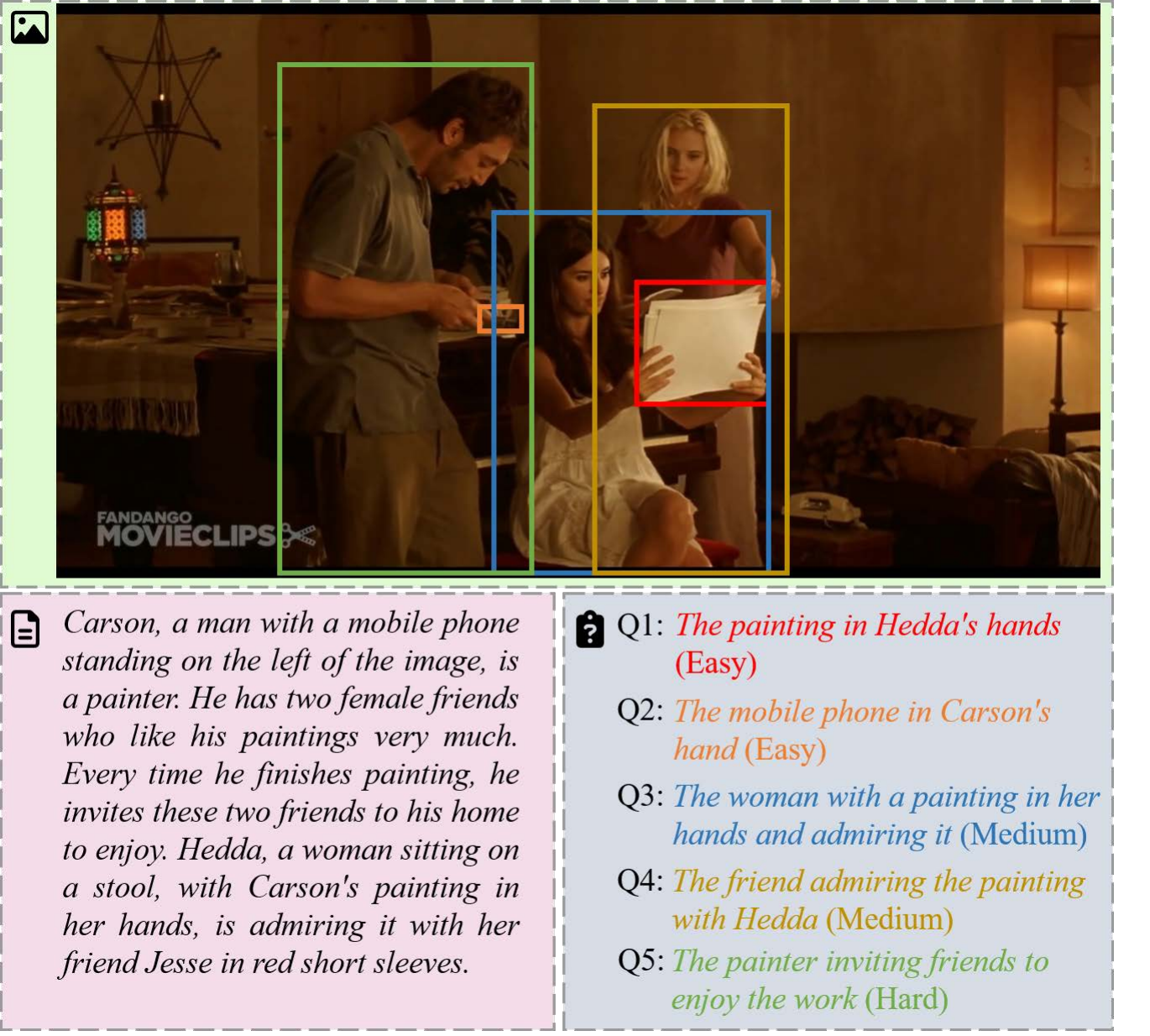}
\caption{An example of the test set. Each query is assigned a difficulty level according to the significance of knowledge for the query. Specifically, Q1 (``\textit{The painting}'') and Q2 (``\textit{The mobile phone}'') are visually-obvious and thus classified into \textbf{Easy} level. Q3 and Q4 are knowledge-relevant but provide some visual clues (``\textit{a painting in her hands}'' and ``\textit{admiring the painting}''), rated as \textbf{Medium} level. Q5 relies heavily on the knowledge and deserves a \textbf{Hard} label.}
\label{fig:example8}
\end{figure*}

\begin{figure*}[t]
\centering
\includegraphics[width=0.95\textwidth, trim=0 0 0 0]{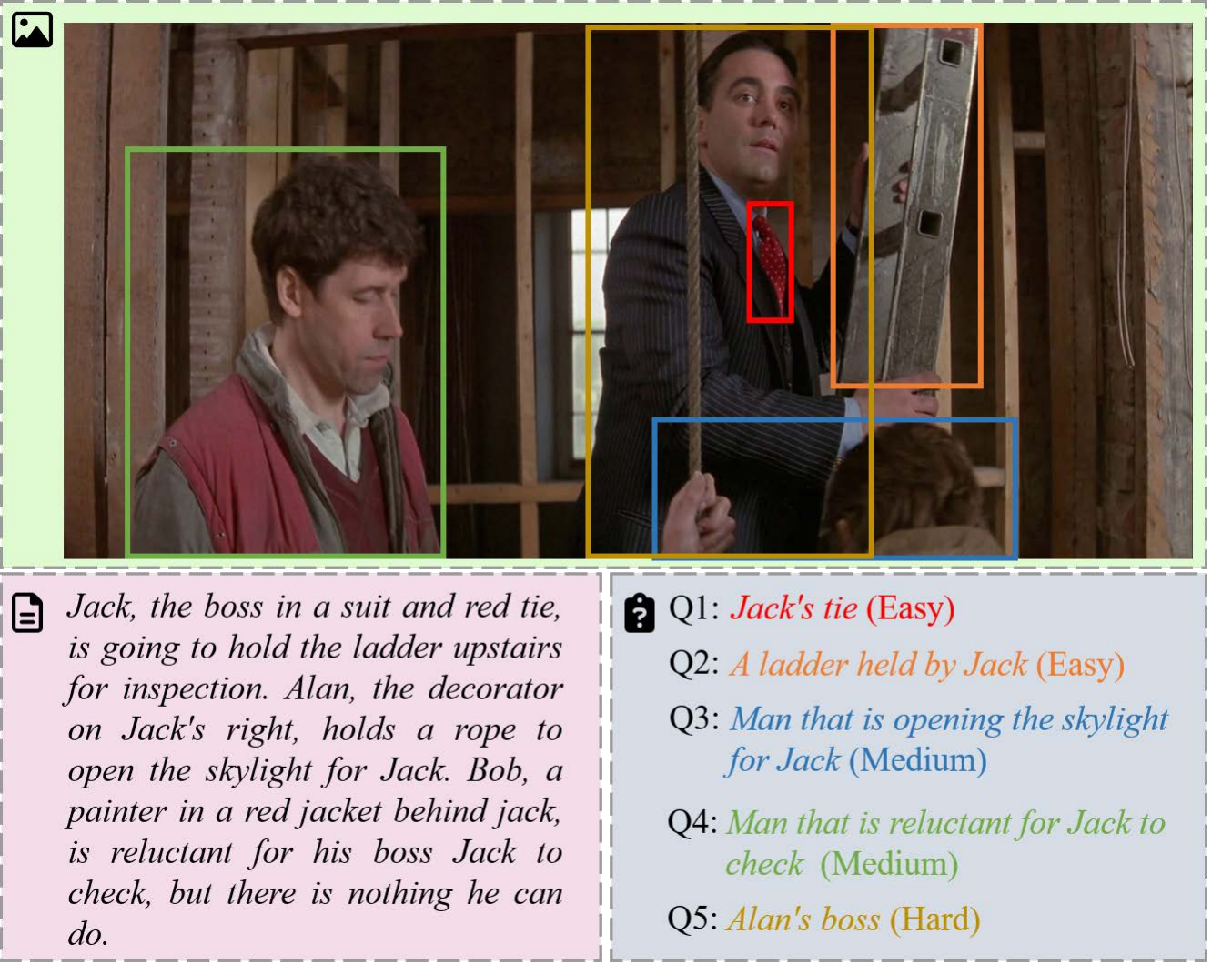}
\caption{An example of the test set. Each query is assigned a difficulty level according to the significance of knowledge for the query. Specifically, Q1 (``\textit{tie}'') and Q2 (``\textit{A ladder}'') are visually-obvious and thus classified into \textbf{Easy} level. Q3 and Q4 are knowledge-relevant but provide some visual clues (``\textit{opening the skylight}'' and ``\textit{reluctant}''), rated as \textbf{Medium} level. Q5 relies heavily on the knowledge and deserves a \textbf{Hard} label.}
\label{fig:example9}
\end{figure*}

We show more examples of the proposed SK-VG dataset.
For ease of reading, we make a figure for each example and show them in Figure~\ref{fig:example2}-\ref{fig:example9}.

% ****************** Figure 1 ******************
% ****************** Figure 1 ******************

% ****************** Figure 2 ******************
% ****************** Figure 2 ******************

% ****************** Figure 3 ******************
% ****************** Figure 3 ******************

% ****************** Figure 4 ******************
% ****************** Figure 4 ******************

% ****************** Figure 5 ******************
% ****************** Figure 5 ******************

% ****************** Figure 6 ******************
% ****************** Figure 6 ******************

% ****************** Figure 7 ******************
% ****************** Figure 7 ******************

% ****************** Figure 8 ******************
% ****************** Figure 8 ******************

% ****************** Figure 9 *****************
% ****************** Figure 9 *****************

\ifarxiv \clearpage  \fi

\end{document}